\documentclass[preprint,12pt,authoryear]{elsarticle}

\usepackage{amssymb}
\usepackage[authoryear]{natbib}
\usepackage{subcaption}

\usepackage{tikz}
\usepackage{float}
\usepackage[edges]{forest}
\usepackage[T1]{fontenc}
\usepackage{lmodern}   % <- MUY IMPORTANTE
\usepackage{multirow}
\usepackage{booktabs}
\usepackage{amsmath}
\usepackage{url}
\usepackage{gensymb}
\usepackage{graphicx}
\usepackage{verbatim}
\usepackage{tabularx} % Add this in the preamble if not already included
\usepackage{algorithm}
\usepackage{algorithmicx}
\usepackage{algpseudocode}
\usepackage{minted}
\usepackage{listings}
\usepackage[utf8]{inputenc}
\RequirePackage{stfloats}
\usepackage{xurl}
\usepackage[colorlinks=true, linkcolor=blue, citecolor=blue, urlcolor=blue, hidelinks]{hyperref} 
\usepackage{adjustbox}

\def\tsc#1{\csdef{#1}{\textsc{\lowercase{#1}}\xspace}}
\tsc{WGM}
\tsc{QE}
\begin{document}

\begin{frontmatter}

% ------------------------------------------------
% TITLE
% ------------------------------------------------

\title{GreenSeg: Ground Segmentation Algorithm for Agricultural Robots in Mediterranean Greenhouses using RGB-D Point Clouds}

% ------------------------------------------------
% AUTHORS
% ------------------------------------------------

\author[UAL-Inf]{Fernando Cañadas-Aránega}\corref{corr}\ead{fernando.ca@ual.es}
\cortext[corr]{Corresponding author}

\author[UAL-Inf]{José C. Moreno}\ead{jcmoreno@ual.es}

\author[UAL-Eng]{José L. Blanco-Claraco}\ead{jlblanco@ual.es}

% ------------------------------------------------
% AFFILIATIONS
% ------------------------------------------------

\affiliation[UAL-Inf]{
    organization={Department of Informatics, CIESOL, ceiA3, Universidad de Almería},
    addressline={Ctra. Sacramento s/n},
    city={Almería},
    postcode={04120},
    country={Spain}
}

\affiliation[UAL-Eng]{
    organization={Department of Engineering, CIESOL, ceiA3, Universidad de Almería},
    addressline={Ctra. Sacramento s/n},
    city={Almería},
    postcode={04120},
    country={Spain}
}

% ------------------------------------------------
% ABSTRACT
% ------------------------------------------------

\begin{abstract}
Greenhouse agriculture in the Mediterranean region faces significant automation challenges due to its unique structural and environmental constraints. These environments are characterized by extremely narrow aisles, heterogeneous terrains—ranging from concrete to tilled soil—and severe optical interference caused by polyethylene covers, which induce specular reflections and 'ghost points' in depth sensors. While autonomous navigation is essential for digitizing agricultural tasks, traditional solutions often rely on expensive 3D LiDAR systems that are economically unscalable for most facilities. To address this, this paper presents GreenSeg, a robust perception framework for autonomous navigation using RGB-D sensing. The proposed method introduces a dual-layer validation strategy: a robust global plane fitting combined with a surface curvature filter for terrain adaptability, and a seed-point-based Region Growing constraint to ensure the spatial continuity of the navigable plane. Experimental validation was conducted using the AGRICOBIOT I platform across four diurnal scenarios with varying solar elevations. The results show that GreenSeg consistently outperforms benchmark segmentation methods, achieving peak improvements of 11.58\% in mean Recall and 19.24\% in mIoU during critical rotational maneuvers at the end of corridors. These findings confirm that the proposed algorithm enables stable and safe autonomous navigation in unstructured, dynamic agricultural environments that are subject to budget constraints and sensitive to lighting conditions.
\end{abstract}

% ------------------------------------------------
% KEYWORDS
% ------------------------------------------------


\end{frontmatter}

% Main text
\section{Introduction} \label{1. Intrducciotion}

Over the past few decades, greenhouse agriculture has established itself as a fundamental strategy for increasing food production and optimizing the management of water and energy resources \citep{duckett2018agricultural}. These advancements are driven by the progressive integration of automated systems and advanced control mechanisms into critical processes, such as microclimate management. Within the realm of agricultural automation, mobile robots have emerged as essential agents in the transition toward a digitized agricultural framework, undertaking transport, inspection, and assistance tasks traditionally deemed labor-intensive or hazardous \citep{vougioukas2019agricultural,moreno2024feasibility}. However, despite being relatively structured environments, greenhouses impose unique challenges for autonomous mobility that diverge significantly from those encountered in standard urban or industrial settings \citep{canadasaranega2026benchmark}.

Mediterranean-type greenhouses, which account for 92\% of the global protected cultivation area, feature a structural configuration designed to maximize crop space at the expense of maneuverability. These environments are characterized by extremely narrow aisles (ranging from 90 to 110 cm), which drastically constrains the kinematic margin of maneuver and demands centimeter-level precision for safe navigation \citep{canadas2024agricobiot,lytridis2021overview}. Furthermore, the terrain exhibits complex, non-linear dynamics; the ground composition can alternate abruptly among compacted sand, concrete, gravel, and tilled soil. This heterogeneity, often compounded by crop residues and variable slopes (typically ranging from 1\% to 2\%), significantly increases the uncertainty in the robot's pose estimation, as we have already discussed in previous papers \citep{canadas2024pid, Canadasifac2026}. A critical bottleneck for robotic perception in such environments is the optical interference generated by polyethylene covers \citep{firkat2023fgseg}. The white plastic materials and metallic frameworks induce specular reflections and glare phenomena that saturate depth sensors, generating "ghost points" or multipath reflection artifacts below the actual ground level. To ensure safe autonomous navigation, it is imperative to deploy robust ground segmentation algorithms capable of filtering these virtual artifacts and dynamically adapting to the irregular topography in real time \citep{lee2022patchwork}.

In contrast to conventional navigation conventional navigation frameworks that rely on expensive omnidirectional 3D LiDAR sensors, the use of cameras as an alternative has proved particularly useful in recent years \citep{song2023navigation}. The implementation of adaptive segmentation algorithms on such sensors facilitates the extraction of dense point clouds for accurate navigable space identification, thereby significantly reducing technological deployment costs in agricultural facilities. By integrating specialized modules for multipath noise suppression and dynamic ground probability estimation, the proposed system ensures highly efficient segmentation, maintaining robustness even in the presence of severe optical saturation and the extreme terrain variability typical of greenhouse infrastructures \citep{luo2022field}.

The main objective of this research is to validate a robust segmentation algorithm using the AGRICOBIOT I mobile platform \citep{lopez2023navigation}, relying exclusively on low-cost depth vision systems, thereby significantly improving the implementation of autonomous navigation in dynamic and complex environments. This study proposes a novel dual-layer validation architecture for the camera-extracted point cloud: a surface curvature filter integrated with mathematical plane fitting to handle terrain irregularities, and a Region Growing connectivity constraint to ensure the spatial integrity of the navigable plane. System performance was rigorously evaluated across complex topological zones—including narrow crop rows, central corridors, and transitions between heterogeneous materials (concrete, gravel, and tilled soil)—while assessing its robustness against extreme diurnal lighting variations and platform kinematics. Experimental results demonstrate that this new approach, termed GreenSeg, achieves peak improvements of 11.58\% in mean Recall and 19.24\% in mIoU compared to the baseline algorithm during critical turning maneuvers. These findings yield superior segmentation in dynamic environments characterized by severe shadows, depth discontinuities, and polyethylene-induced optical interference. Ultimately, this research confirms that precise and safe mobility can be achieved within Mediterranean greenhouse infrastructures utilizing affordable sensory suites, effectively overcoming traditional constraints associated with sensor ''ghosting`` and complex terrain irregularities without the need for high-cost LiDAR systems.

The principal challenges and problematics that this paper aims to address are summarized as follows:

\begin{itemize}
    \item \textit{Structural Morphology of Mediterranean Greenhouses}: The unstructured and dense growth patterns of the crops frequently occlude the ground, complicating the distinction between the navigable plane and the lower canopy. This results in highly constrained aisles (typically 90 to 110 cm wide), which drastically reduces the robot's kinematic margin and necessitates centimeter-level precision to avoid crop collisions, a requirement largely overlooked in the current literature. Furthermore, the white polyethylene covers induce massive specular reflection noise. This phenomenon generates "ghost points" or virtual point clouds below the ground level, confusing depth sensors and severely hindering precise segmentation.
    
    \item \textit{Terrain Morphology of Mediterranean Greenhouses}: The ground in these environments is highly heterogeneous and exhibits complex non-linear dynamics. Surface composition can transition abruptly among compacted sand, concrete, gravel, or tilled soil, frequently interspersed with obstacles such as irrigation hoses and harvest residues. Additionally, unlike flat industrial environments, greenhouses typically feature variable slopes (ranging from 1\% to 2\%) and local elevations, which significantly increase the uncertainty in the robot's pose estimation and dynamic stability.
    
    \item \textit{Limitations of Traditional Navigation Frameworks}: Within greenhouse infrastructures, Global Positioning System (GPS)/ Global Navigation Satellite System (GNSS) signal reception is severely attenuated or completely denied due to the metallic framework and dense vegetation canopy. This forces an exclusive reliance on local exteroceptive sensors for autonomous navigation. Consequently, there is a critical need to develop cost-effective perception alternatives (such as RGB-D cameras) that offer a scalable advantage over existing robust navigation systems that depend heavily on expensive 3D LiDARs.

    \item \textit{Environmental Conditions}: The perceptual system must maintain its segmentation efficacy throughout the diurnal cycle, exhibiting resilience against continuous variations in solar elevation and azimuth. These dynamic lighting conditions directly exacerbate the glare on polyethylene surfaces and compromise the reliability of optical sensors.
    
\end{itemize}

The remainder of this paper is organized as follows. Section~\ref{sec: RW} reviews the state of the art in ground segmentation and environmental perception for agricultural robotics. Section~\ref{sec: MM} describes the experimental site, the AGRICOBIOT~I robotic platform, and the baseline perception pipeline. Section~\ref{sec: camera} presents the proposed GreenSeg algorithm and the YOLOx-based farmer detection module. Section~\ref{sec: result} reports the experimental results under the four navigation trajectories and varying solar conditions. Finally, Section~\ref{sec: conclusion} summarizes the main conclusions and outlines directions for future work.

\section{Related works} \label{sec: RW}

Ground segmentation and environmental perception constitute the fundamental pillars of autonomous navigation in agriculture. Over the years, the dominant paradigm has evolved from conventional image processing techniques to advanced deep learning models tailored for unstructured scenes \citep{huang2021fast}.

\subsection{Traditional Image Processing Techniques}

Traditional ground segmentation techniques predominantly rely on the physical and geometric characteristics of point clouds \citep{mijit2023lr}. Among the most widely adopted methods is plane fitting via RANSAC (Random Sample Consensus), which seeks the geometric plane that maximizes the number of inliers. However, these models frequently suffer from under-segmentation in non-planar terrains, such as greenhouses characterized by variable slopes and surface irregularities \citep{lee2022patchwork}. To enhance terrain representation, several studies have proposed the use of elevation maps, which discretize the space into a 2.5D grid and facilitate obstacle identification through height variance analysis. Nevertheless, these methods are prone to failure in the presence of dense vegetation or overhanging structures, often yielding false positive detections \citep{paigwar2020gndnet}.

In this context, algorithms such as Patchwork and its successor, Patchwork++, have represented a significant breakthrough by introducing region-based segmentation strategies and adaptive terrain analysis, achieving a robust balance between accuracy and computational efficiency in complex environments \citep{muthukumaraswamy2025integrating}. Similarly, techniques like Cloth Simulation Filtering (CSF) model the ground as a deformable surface, yielding satisfactory results in natural scenes; however, they remain highly sensitive to noise and require extensive manual parameter tuning \citep{yang2025planesegnet}. Other probabilistic approaches, such as those based on Markov Random Fields (MRF), have successfully modeled the spatial coherence of the terrain, but their prohibitive computational cost restricts their applicability under real-time constraints \citep{galin2019review}. Overall, these traditional methods struggle to generalize across highly dynamic and visually complex environments like greenhouses. These geometric assumptions often collapse in Mediterranean greenhouses, where the high-reflectivity polyethylene covers generate multipath artifacts that RANSAC-based methods mistakenly incorporate into the ground plane, leading to critical errors in robot pose estimation.

\subsection{Deep Learning and Machine Learning Vision Model}

The advent of deep learning has overcome the inherent limitations of hand-crafted descriptors through the automated learning of hierarchical representations \citep{lee2022patchwork}. Pioneering architectures such as PointNet and PointNet++ introduced the direct processing of unordered point clouds, achieving remarkable results in object classification and semantic segmentation \citep{luo2022field}. Nonetheless, these networks typically struggle to model local contextual geometries in large-scale, high-density scenes \citep{yang2025planesegnet}.

To address these shortcomings, more advanced architectures like KPConv, RandLA-Net, and SparseConvNet have been developed. These frameworks introduce efficient spatial convolutions and attention mechanisms to enhance the representation of geometric structures \citep{xie2020segcloud, paigwar2020gndnet}. RandLA-Net, in particular, has demonstrated outstanding real-time performance owing to its random sampling strategy and hierarchical feature aggregation. Another highly relevant research avenue is deep supervised learning specifically designed for ground/non-ground classification, exemplified by GndNet. This network learns terrain distributions directly from LiDAR data, effectively bypassing the constraints of classical geometric methods \citep{lentsch2026terraseg}.

To tackle computational efficiency, Sparse Convolutional Networks (SCNs) and voxel-based methods have been engineered to execute convolution operations exclusively in non-empty regions, thereby drastically reducing memory consumption \citep{yang2025planesegnet}. Models like RandLA-Net have set benchmarks in efficiency by employing random sampling alongside lightweight attention mechanisms to process millions of points in real time \citep{xie2020segcloud}. In recent investigations, the integration of Plane Attention modules has enabled the capture of critical vertical geometric variations at ground-plant interfaces, substantially improving accuracy in boundary regions where structural limits are ambiguous. Furthermore, multi-view strategies such as SnapNet project the point cloud into multiple 2D snapshots to leverage the maturity of image-based CNNs, subsequently performing a back-projection into the 3D space for the final semantic labeling \citep{boulch2018snapnet}. While state-of-the-art deep learning architectures offer high semantic accuracy, their deployment on lightweight agricultural platforms remains constrained by prohibitive computational demands and a lack of training datasets that specifically account for the 'ghost point' phenomena prevalent in plastic-covered environments

\subsection{Semantic Segmentation in Precision Agriculture}

In recent years, ground and vegetation segmentation within agricultural environments has gravitated toward multi-modal approaches combining RGB, RGB-D, and LiDAR sensors \cite{canadas2024greenbot}. These strategies aim to enhance perceptual robustness under real-world conditions characterized by extreme visual and structural variability \citep{qian2014ncc,lentsch2026terraseg}. In the computer vision domain, several studies have deployed semantic segmentation networks such as U-Net, DeepLabv3+, and SegNet to differentiate among soil, crop, and weed classes using RGB imagery. For instance, Milioto et al. \citep{milioto2018real} proposed a CNN-based approach for semantic segmentation in precision agriculture, demonstrating high accuracy in structured crop scenarios. Similarly, more recent works have incorporated RGB-D data to refine depth estimation and improve the separation between ground and vegetation planes, enabling superior discrimination in scenes with partial occlusions.

Conversely, the utilization of LiDAR sensors has gained substantial traction due to their ability to capture highly accurate geometric information. In this context, Paigwar et al. \citep{paigwar2020gndnet} introduced GndNet, a neural network specifically tailored for ground segmentation in LiDAR point clouds, outperforming traditional geometric methods in complex environments. Similarly, Sa et al. \citep{sa2016deepfruits} developed deep learning-based approaches for semantic terrain classification in agricultural settings by integrating both geometric and contextual features. Moreover, works such as Kang et al. \citep{kang2023semantic} have explored the fusion of LiDAR data and RGB cameras to enhance crop and ground segmentation, exploiting both spectral and structural modalities. These multi-modal frameworks have proven particularly effective in scenarios subjected to highly variable illumination, where vision-only methods frequently degrade.

Within greenhouse environments, where structural density is higher and lighting conditions are significantly more volatile, recent studies have tackled segmentation through hybrid methodologies \cite{he2022automated}. Despite these advancements, the majority of existing frameworks exhibit notable limitations concerning generalization capabilities, computational efficiency, or an over-reliance on controlled conditions. Specifically, purely vision-based methods remain highly susceptible to illumination shifts, whereas LiDAR-exclusive approaches often struggle to discriminate between the ground and low-lying vegetation \citep{de2024deep}. Consequently, there remains a pressing need to develop robust, real-time segmentation algorithms capable of adapting to the idiosyncratic conditions of greenhouses, where heterogeneous surfaces, complex structures, and multiple dynamic agents invariably coexist. Despite the robustness of multi-modal frameworks, existing agricultural segmentation solutions frequently degrade under the dynamic diurnal lighting cycles of greenhouses, where extreme glare saturates RGB channels and LiDAR-based discrimination between low-lying crops and heterogeneous soil remains an open challenge. This work aligns with this imperative, proposing an approach specifically geared toward highly reliable ground segmentation for autonomous navigation in real-world agricultural settings.

\section{Materials and methods} \label{sec: MM}

This section details the experimental framework and methodological procedures employed to validate the proposed cost-effective RGB-D ground segmentation system within Mediterranean greenhouse environments. First, the experimental site located at the AgroConnect facilities is described; this setting features heterogeneous terrain (comprising sand, concrete, and gravel) alongside polyethylene plastic covers that induce significant optical artifacts. Subsequently, the AGRICOBIOT II mobile platform is introduced, a differential-drive robot specifically engineered for payload transportation within these confined spaces. The methodological core of this work lies in the development of a perception framework based on the Patchwork++ 3D point cloud segmentation algorithm, specifically adapted for integration with the Intel® RealSense™ D435 camera.

\subsection{Materials}

A description of the materials used is provided, based on the description of the greenhouse and the actual robot model.

\subsubsection{AgroConnect greenhouse facilities}

\begin{figure}[!ht]
    \centering
    \subfloat[Greenhouse outdoor \label{fig:sub1}]{
        \includegraphics[width=9cm]{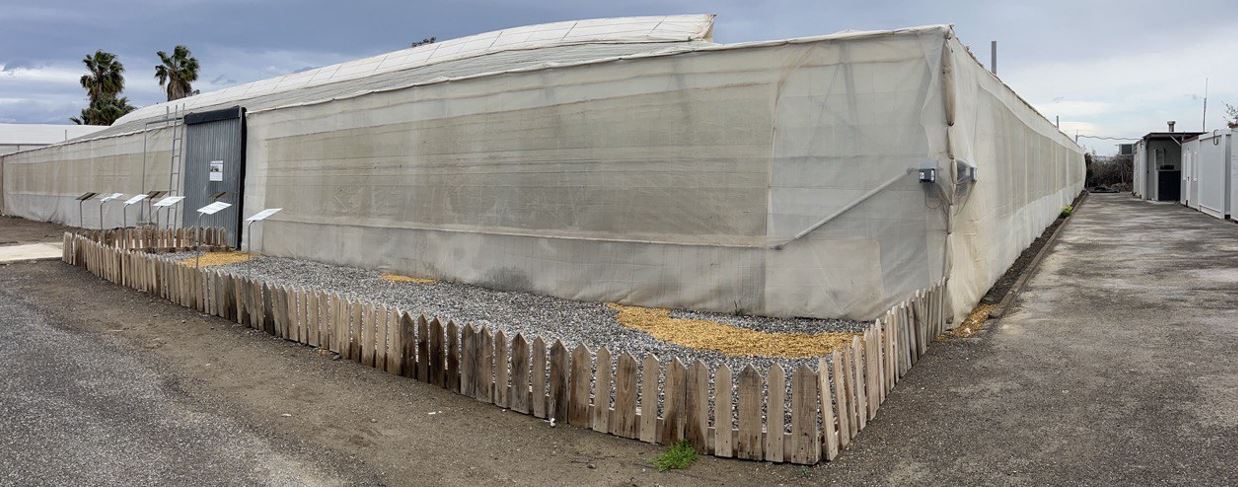}
    }
    \\
    \subfloat[Greenhouse indoor \label{fig:sub2}]{
        \includegraphics[width=9cm]{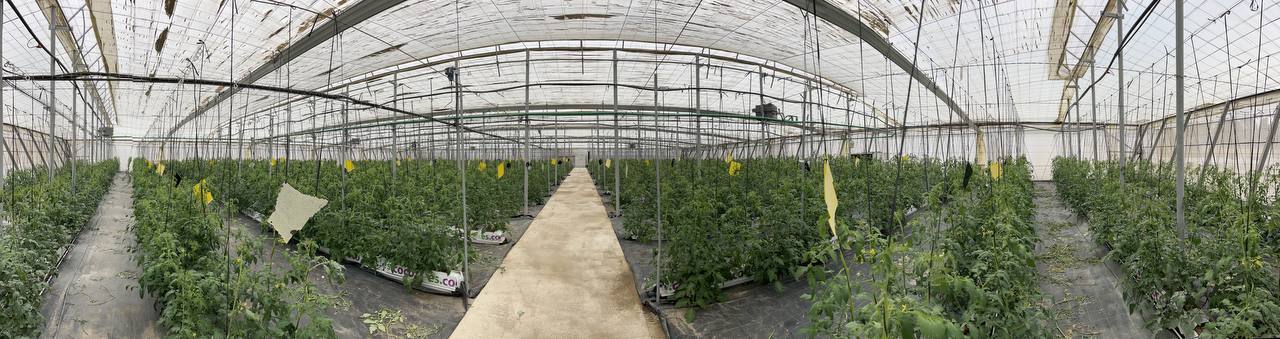}
    }
    \caption{AgroConnect greenhouse}
    \label{fig:RobotAGI2}
\end{figure}

The experimental validation was conducted at the AgroConnect research facilities (Almería, Spain; $36^{\circ}50' \text{ N}, 2^{\circ}24' \text{ W}$), a site co-funded by the Spanish Ministry of Science, Innovation, and Universities and the European Regional Development Fund (FEDER) (Figure \ref{fig:sub1}). Situated at an elevation of 3 m above sea level, the terrain exhibits a consistent 1\% northern gradient. The testbed comprises a 1,850 m\textsuperscript{2} Mediterranean-style greenhouse (traditionally classified as a \textit{raspa y amagado} structure) featuring a robust steel framework and high-density polyethylene cladding. Topologically, the facility is structured around a 2 m wide central corridor serving as the primary navigational artery. This axis connects to eleven lateral aisles on each side; the northern aisles measure 12.5 m in length, while the southern aisles extend to 22.5 m, both constrained to a width of 2 m without taking plants into account (Figure \ref{fig:sub2}).

\subsubsection{AGRICOBIOT I robot}

\begin{figure}[!ht]
    \centering
    \subfloat[AGRICOBIOT I in a corridor with high plants\label{fig:AGIIm2}]{
        \includegraphics[width=0.465\linewidth]{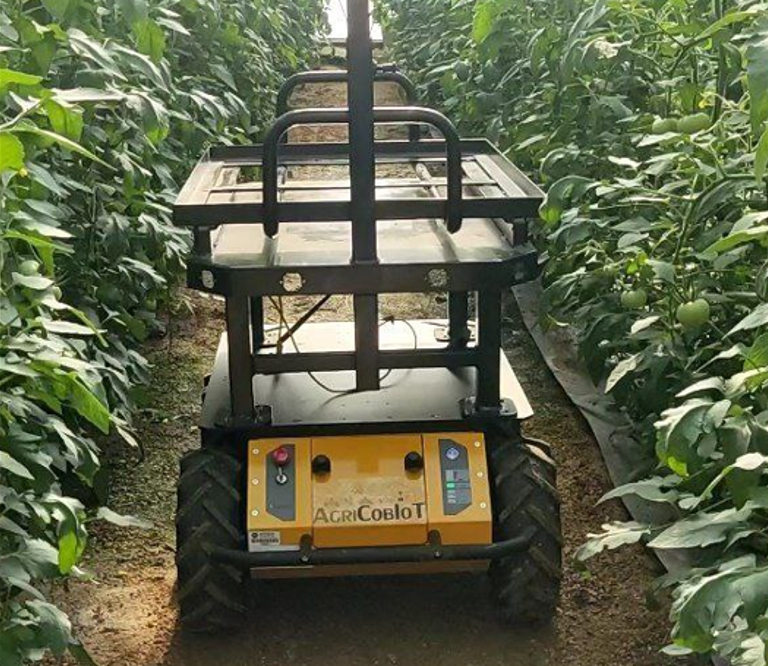}
    }
    %\hfill
    \subfloat[AGRICOBIOT I in a corridor with low plants\label{fig:AGIIr2}]{
        \includegraphics[width=0.4\linewidth]{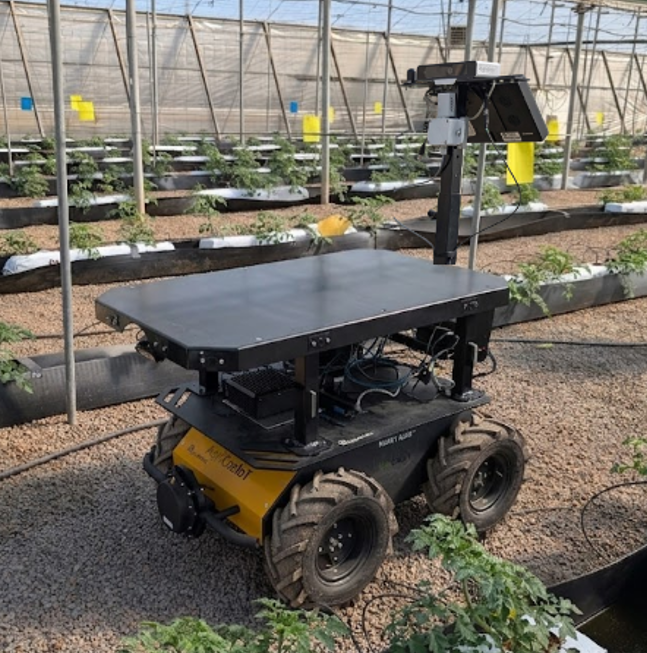}
    }
    \caption{AGRICOBIOT I robot}
    \label{fig:RobotAGI}
\end{figure}

The experimental validation involves the deployment of the AGRICOBIOT I \citep{lopez2023navigation} mobile platform, seamlessly integrated into a fog-based architecture to facilitate real-time data exchange across the facility's broader operational network. Built upon a commercial Husky chassis by Clearpath Robotics\textsuperscript{\textregistered}, this platform employs a differential drive system capable of zero-radius turning, a kinematic requisite for navigating the severely constrained aisles of greenhouse environments. To optimize agricultural utility, the robot is outfitted with a custom-engineered superstructure featuring four central posts and a removable, rail-guided implement; this design facilitates the transport of standardized harvest boxes at an ergonomic height while maximizing the available payload space for the perceptual suite. Figure \ref{fig:RobotAGI} shows various views of the AGRICOBIOT I robot.

To execute the experimental validation, the robotic platform is equipped with the following heterogeneous computational and sensory suite:

\begin{itemize}
    \item \textit{HITTSON Unit Control}: This high-performance industrial onboard computer serves as the central processing hub for the robotic agent. It integrates a Wi-Fi 6 (802.11ax) wireless module, establishing high-bandwidth, low-latency communication with the local greenhouse network infrastructure.
    
    \item \textit{Intel\textsuperscript{\tiny\textregistered} RealSense\textsuperscript{\tiny\texttrademark} D435 Camera}: An active stereo depth camera deployed for core visual perception and identification tasks. Within this architecture, it acquires the dense RGB-D data essential for the perceptual pipeline, enabling the system to extract precise spatial coordinates of dynamic agents (e.g., human operators and peer robots) while serving as the primary sensory input for the proposed ground segmentation framework.

\end{itemize}

\subsection{Methods}

This section describes the methods and algorithms used in the development of ground segmentation systems.

\subsubsection{Robot Operating System 2 (ROS 2)}

The robotic software architecture is fundamentally anchored in ROS 2 Humble Hawksbill, serving as the decentralized middleware framework for hardware abstraction and high-level task orchestration. Leveraging the Data Distribution Service (DDS) standard, this architecture implements rigorous Quality of Service (QoS) policies, which are critical for ensuring deterministic and reliable inter-process communication amidst the severe network degradation typical of greenhouse environments \citep{macenski2022ros2}. Within this multi-agent paradigm, ROS 2 guarantees seamless interoperability via standardized messaging protocols, enabling heterogeneous platforms to construct and share a unified world model. To ensure scalability and mitigate data contention, each robotic unit operates within a strictly defined namespace. This architectural isolation permits a dedicated MQTT bridge to seamlessly aggregate local perceptual telemetry into a centralized fleet management layer devoid of cross-talk. Furthermore, the framework integrates the \texttt{tf2} transformation library to maintain rigorous spatial synchronization across all agents, dynamically projecting localized semantic detections (e.g., via YOLOx) into a globally consistent coordinate frame. Ultimately, the confluence of decentralized DDS communication and real-time spatial synchronization elevates individual autonomous nodes into a cohesive, perception-aware fleet. This synergy is fundamental for executing the complex cooperative navigation and collaborative human-robot interactions demanded by high-density agricultural settings \citep{macenski2020marathon}.

\subsubsection{Base Groundfloor Segmentation Framework}

The perception pipeline is built upon a ground-floor segmentation algorithm developed by Intel\textsuperscript{\textregistered} and distributed as part of the Robotics~AI Suite~\cite{intelrealsense_groundfloor}. The algorithm, based on Ground Plane Fitting (GPF) method, ingests raw 3-D sensor data produced by either Intel\textsuperscript{\textregistered} RealSense\textsuperscript{\texttrademark} active-stereo depth cameras or rotating 3D LiDAR scanners, and assigns one of four semantic labels to every valid measurement point, thereby partitioning the workspace into a navigable ground manifold, positive collision obstacles, overhanging structures, and sensor noise artefacts. This is achieved by retrieving the depth data published in ROS 2 by the Intel RealSense camera, along with the camera's intrinsic and extrinsic parameters, as provided by the manufacturer.

The algorithm operates on a raw point cloud $\mathcal{P} = \{\mathbf{p}_i\}_{i=1}^{N}$ (where $i, N \in \mathbb{N}^+$) to transform it into the robot's coordinate frame, ensuring that each point $\mathbf{p}_i = (x_i, y_i, z_i)^\top \in \mathbb{R}^3$ is expressed relative to the robot's reference frame (\texttt{base\_link}). Each point captured in the camera's optical frame is transformed via the ROS~2 TF2 tree according to the spatial relationship defined in Equation~\ref{eq:transform}.

\begin{equation}
    \mathbf{p}_i^{\,\text{base}}
        \;=\; T_{\text{cam}}^{\text{base}} \cdot \mathbf{p}_i^{\,\text{cam}}
        \;=\; \mathbf{R}\, \mathbf{p}_i + \mathbf{t},
    \label{eq:transform}
\end{equation}

\noindent where $T_{\text{cam}}^{\text{base}} \in SE(3)$ denotes the rigid body transformation (comprising a rotation matrix $\mathbf{R} \in SO(3)$ and a translation vector $\mathbf{t} \in \mathbb{R}^3$) between the sensor and the \texttt{base\_link} frame. This transformation is obtained in real-time by querying the ROS~2 TF2 transform tree at the specific timestamp of each frame. Subsequently, a secondary pre-filtering stage based on a minimum distance threshold is implemented to discard points in the immediate vicinity of the robot. This region is typically prone to self-specular reflections, occlusion shadows, and sensor saturation artifacts. To mitigate these effects, a radial distance threshold $d_{\min} \in \mathbb{R}^+$ is applied within the $xy$ plane of the robot's reference frame, as formulated in Equation~\ref{eq:prefilter}.

\begin{equation}
    \mathcal{P}_{\text{valid}}
        = \left\{
            \mathbf{p}_i \in \mathcal{P}
            \;\middle|\;
            \sqrt{x_i^2 + y_i^2} \;\geq\; d_{\min}
          \right\}.
    \label{eq:prefilter}
\end{equation}

The subsequent phase focuses on identifying the dominant traversable surface through a Ground Fitting algorithm, which employs planar adjustment heuristics. Geometrically, the ground plane is parameterized by its unit normal vector $\mathbf{n} = (n_x, n_y, n_z)^\top\in \mathbb{R}^3$ ($\|\mathbf{n}\|=1$) and a scalar offset $d \in \mathbb{R}$, such that any point $p$ belonging to the plane satisfies Equation~\ref{eq:plane}.
\begin{equation}
    \mathbf{n}^\top p + d \;=\; 0.
    \label{eq:plane}
\end{equation}

The estimation process seeks the dominant horizontal manifold. To enforce this, a horizontal prior is implemented by constraining the normal vector to be approximately vertical, as defined in

\begin{equation}
    n_z \;\gg\; n_x,\; n_y
    \qquad \Longrightarrow \qquad
    \mathbf{n} \;\approx\; (0,\; 0,\; 1)^\top.
    \label{eq:horizontalprior}
\end{equation}

The signed perpendicular distance $\delta_i\in \mathbb{R}$ from each candidate point to the estimated plane is given by

\begin{equation}
    \delta_i \;=\; \hat{\mathbf{n}}^\top \mathbf{p}_i + \hat{d}.
    \label{eq:signeddist}
\end{equation}

Utilizing the estimated plane parameters $(\hat{\mathbf{n}}, \hat{d})\in \mathbb{R}^4$ and a set of predefined heuristic thresholds, each point $\mathbf{p}_i \in \mathcal{P}_{\text{valid}}$ is assigned a semantic label $\ell_i$ according to the following classification logic (equation \ref{eq:classification}).
\begin{equation}
    \ell_i \;=\;
    \begin{cases}
        \textit{ground}
            & \text{if } |\delta_i| \leq h_{\text{ground}}
              \;\text{and}\; |\theta| \leq \theta_{\max}, \\[4pt]
        \textit{obstacle}
            & \text{if } h_{\text{ground}} < z_i
              \leq H_{\text{robot}}, \\[4pt]
        \textit{above}
            & \text{if } z_i > H_{\text{robot}}, \\[4pt]
        \textit{noise}
            & \text{otherwise,}
    \end{cases}
    \label{eq:classification}
\end{equation}

\noindent where $\theta \;=\; \arccos\!\left( \hat{n}_z \right)\in [-90\degree, 90 \degree]$  and $\theta_{max}<90\degree$. The parameters represent the following physical constraints:

\begin{itemize}
    \item $h_{\text{ground}}\in \mathbb{R}^+$: The maximum perpendicular tolerance allowed for a point to be categorized as traversable ground.

    \item $\theta_{\max}\in \mathbb{R^+}$: The maximum permissible inclination of the fitted plane relative to the horizontal.

    \item $H_{\text{robot}}\in \mathbb{R^+}$: The physical height of the robot, acting as the decision boundary between the \textit{obstacle} and \textit{above} classes.
\end{itemize}

The plane parameters $(\hat{\mathbf{n}}, \hat{d})$ are refined by optimizing the fit over the initial inlier set $\mathcal{P}_{\text{ground}}^{(0)} = \{\mathbf{p}_i \mid |\delta_i| \leq h_{\text{ground}}\}$ through a constrained least-squares formulation:
\begin{equation}
    (\hat{\mathbf{n}},\;\hat{d})
        \;=\; \underset{\mathbf{n},\,d}{\arg\min}
              \sum_{\mathbf{p}_i\,\in\,\mathcal{P}_{\text{ground}}^{(0)}}
              \!\!\!\!\left(
                  \mathbf{n}^\top \mathbf{p}_i + d
              \right)^{\!2}
              \qquad
              \text{with} \quad \|\mathbf{n}\| = 1.
    \label{eq:planefit}
\end{equation}

This optimization admits a closed-form solution: $\hat{\mathbf{n}}$ corresponds to the eigenvector associated with the smallest eigenvalue of the $\textbf{C}_i$, satisfying $\mathcal{P}_{\text{ground}}^{(0)}$, while the offset is determined by $\hat{d} = -\hat{\mathbf{n}}^\top \bar{p}$, where $\bar{p}$ denotes the centroid of the inlier set. Finally, the framework publishes two distinct ROS~2 \texttt{sensor\_msgs/PointCloud2} topics. The first constitutes the complete semantically labeled cloud:

\begin{equation}
    \mathcal{P}_{\text{ground}}
        \;=\; \bigl\{(\mathbf{p}_i,\,\ell_i)\bigr\}_{i=1}^{|\mathcal{P}_{\text{valid}}|}.
    \label{eq:labeled}
\end{equation}

The second is the dedicated obstacle cloud, which is directly integrated into the navigation stack's local cost-map:

\begin{equation}
    \mathcal{P}_{\text{obs}}
        \;=\; \left\{
                \mathbf{p}_i \in \mathcal{P}_{\text{valid}}
                \;\middle|\;
                \ell_i = \textit{obstacle}
              \right\}.
    \label{eq:obstacles}
\end{equation}

\begin{figure}[!ht]
    \centering
    \includegraphics[width=0.75\linewidth]{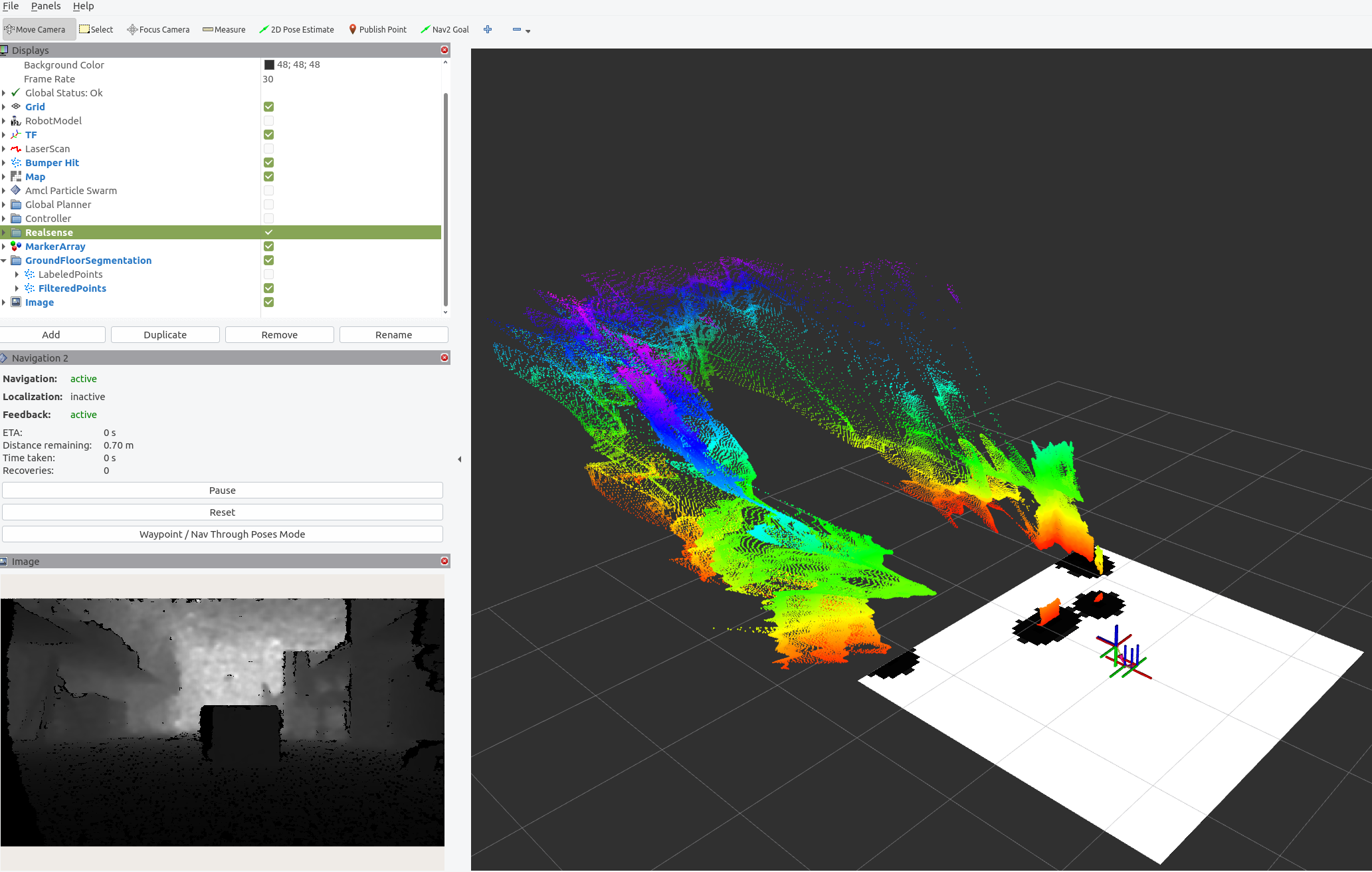}
    \caption{Example of how to use the framework with a jet-type color intensity \cite{intelrealsense_groundfloor}}
    \label{fig:Eje_groound}
\end{figure}

These ROS~2 topics provide the essential perceptual feedback required for autonomous navigation and path planning within the greenhouse environment (Figure~\ref{fig:Eje_groound}):

\begin{itemize}
\item \textit{Ground Floor (Navigable Terrain)}: Extracted by identifying the dominant horizontal plane relative to the camera's extrinsics, this category represents the primary traversable surface essential for path planning.
\item \textit{Positive Obstacles}: Dense point clusters located within the robot's immediate trajectory footprint and bounded by its vertical operating height ($Z$-axis), representing direct, imminent collision threats.
\item \textit{Overhanging Structures}: Spatial data points situated strictly above the robot's maximum physical height threshold (e.g., structural greenhouse arcs or high canopy layers). While vital for global mapping, they are mathematically decoupled from immediate collision-checking routines to save computational bandwidth.
\item \textit{Unclassified / Sensor Noise}: Outlier points systematically filtered out due to sensor specular reflections, severe scattering from the polyethylene covers, or multi-path interference.
\end{itemize}

Architecturally, this base algorithm exhibits optimal performance and remarkable computational efficiency when deployed on highly regular, strictly planar topography. Under such structured and static operating conditions, the absence of severe surface irregularities or variable gradients allows the plane-fitting heuristics to maintain absolute systemic stability. However, this strict flat-world assumption exposes a significant algorithmic vulnerability when extrapolated to the unstructured, variable terrain inherent to Mediterranean greenhouses. The presence of uneven soil, arbitrary slopes, and agricultural waste often leads to under-segmentation, thereby necessitating the advanced algorithmic refinements and robust adaptations proposed within this study to ensure fail-safe autonomous navigation.

\section{A Ground-Segmentation Approach for Greenhouses} \label{sec: camera}

This section describes the proposed method for floor segmentation, specifically for complex environments such as greenhouses.

\subsection{Robot setup}

\begin{figure}[!ht]
    \centering
    \includegraphics[width=0.8\linewidth]{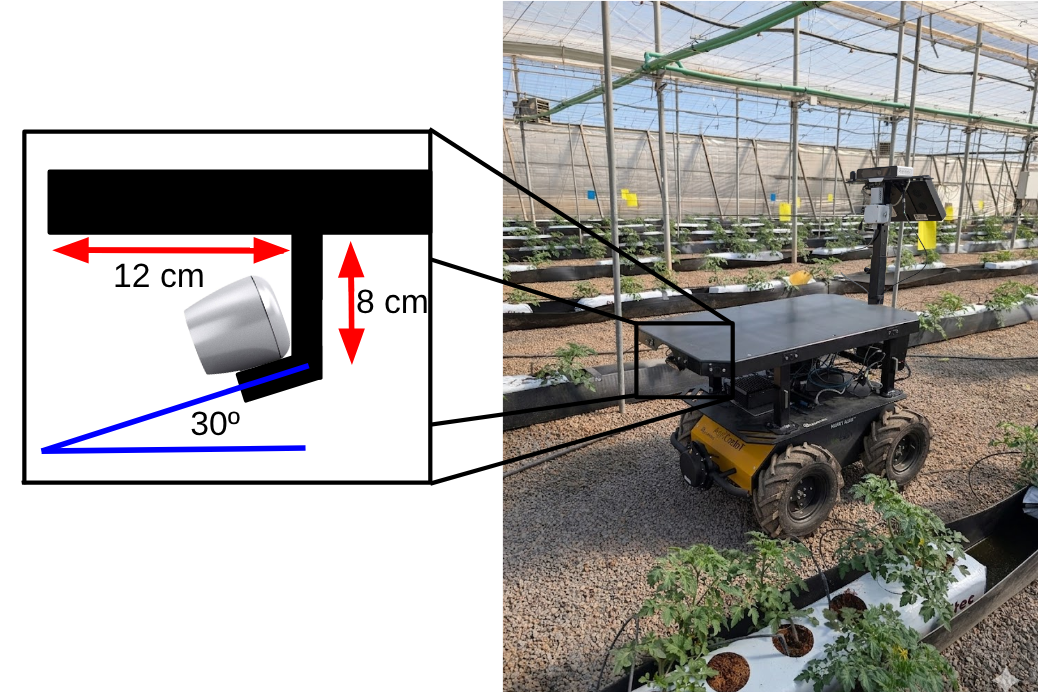}
    \caption{Camera location in robot}
    \label{fig:Camera_posi}
\end{figure}

Initially, the camera was positioned beneath the black transport platform. This recessed positioning is intended to minimise interference from sunlight on the lens, which might otherwise cause unwanted glare and, consequently, false positives in the perception algorithms. In this regard, Figure \ref{fig:Camera_posi} illustrates the camera configuration during one of the tests carried out in a real greenhouse. The sensor was recessed 12~cm to minimize direct sunlight exposure, mounted 8~cm downward to maximize the Field of View (FOV), and tilted at a 30-degree angle to reduce redundant or non-useful visual data.

\subsection{Proposed approach to robust ground identification: GreenSeg}

The base algorithm produces a map $\mathcal{P}_{\text{ground}}$ and an obstacle cloud $\mathcal{P}_{\text{obs}}$, which contains errors. In greenhouse environments, this map may be contaminated by specular reflections from the structure’s polyethylene, highly irregular surfaces, or different types of flooring. The geometric verification stage acts as a second filter, ensuring that only geometrically consistent observations update the temporal model. The aim is to ensure future autonomous navigation in all possible scenarios a robot may encounter in a greenhouse, which will be analyzed and studied in this section. The output of the base algorithm undergoes a second phase for soil classification, where $\mathcal{P}_{\text{ground}}$ and $\mathcal{P}_{\text{obs}}$ are evaluated using the \textit{Region Growing} algorithm \citep{adams1994seeded}, with the new approach proposed in \cite{xu2022consistency}. 

\subsubsection{Local Surface Normal Estimation} \label{subsubsec:rg_normals}

For each measured point $\mathbf{p}_i \in \mathcal{P}_{\text{ground}}^{(0)}$, a local surface normal is estimated from its spherical neighbourhood $\mathcal{N}_i \subset \mathbb{R}^3$ of radius $r\in \mathbb{R}^+$:

\begin{equation}
    \mathcal{N}_i
        = \left\{
            \mathbf{p}_j \in \mathcal{P}_{\text{valid}}
            \;\middle|\;
            \left\| \mathbf{p}_j - \mathbf{p}_i \right\|
            \leq r
          \right\}.
    \label{eq:neighborhood}
\end{equation}

The local covariance matrix $\mathbf{C}_i \in \mathbb{R}^{3\times3}$ of the neighbourhood is

\begin{equation}
    \mathbf{C}_i
        = \frac{1}{|\mathcal{N}_i|}
          \sum_{\mathbf{p}_j \in \mathcal{N}_i}
          \!\!\left(
              \mathbf{p}_j - \bar{\mathbf{p}}_i^{\,\mathcal{N}}
          \right)
          \left(
              \mathbf{p}_j - \bar{\mathbf{p}}_i^{\,\mathcal{N}}
          \right)^{\!\top},
    \label{eq:covmatrix}
\end{equation}

\noindent where $\bar{\mathbf{p}}_i^{\,\mathcal{N}} = \frac{1}{|\mathcal{N}_i|}\sum_j \mathbf{p}_j$ 
is the estimated centroid of $\mathcal{N}_i$. The estimated local normal $\hat{\mathbf{m}}_i\in\mathbb{R}^3$ is defined as the unit eigenvector associated with the smallest eigenvalue $\lambda_{\min}$ of $\mathbf{C}_i$ as $\det(\mathbf{C}_i - \lambda \mathbf{I}) = 0$:

\begin{equation}
    \mathbf{C}_i \,\hat{\mathbf{m}}_i = \lambda_{\min}\,\hat{\mathbf{m}}_i,
    \qquad
    \|\hat{\mathbf{m}}_i\| = 1.
    \label{eq:localnormal}
\end{equation}

Points with $|\mathcal{N}_i| < 30$ (value obtained from \citep{xu2022consistency}) are immediately reclassified as \textit{noise} and excluded from subsequent steps. A point in $\mathcal{P}_{\text{ground}}^{(0)}$ is geometrically consistent with the globally fitted plane $(\hat{\mathbf{n}}, \hat{d})$ (Eq.~\eqref{eq:planefit}) if its local normal is nearly collinear with $\hat{\mathbf{n}}$. The consistency score is defined as the absolute cosine similarity between the two normals:

\begin{equation}
    \rho_i
        = \left|
            \hat{\mathbf{m}}_i^{\top} \hat{\mathbf{n}}
          \right|
        \in [0,\,1].
    \label{eq:normalconsistency}
\end{equation}

A value $\rho_i \approx 1$ indicates that the local surface at ${\mathbf{p}}_i$ is parallel to the globally fitted plane, while $\rho_i \approx 0$ signals a sharp local discontinuity incompatible with flat navigable terrain. The set of candidates passing the normal consistency check is

\begin{equation}
    \mathcal{P}_{\text{ground}}^{\rho}
        = \left\{
            \mathbf{p}_i \in \mathcal{P}_{\text{ground}}^{(0)}
            \;\middle|\;
            \rho_i \geq \rho_{\min}
          \right\},
    \label{eq:normalfilter}
\end{equation}

\noindent where $\rho_{\min} \in (0,\,1]$ is a configurable threshold.
 
\subsubsection{Local Curvature Filtering}
\label{subsubsec:rg_curvature}
 
The surface curvature at point ${\mathbf{p}}_i$ is estimated as the fraction of total neighbourhood variance explained by the normal direction, using the eigenvalues $\lambda_1 \geq \lambda_2 \geq \lambda_3 = \lambda_{\min}$ of $\mathbf{C}_i$:

\begin{equation}
    \kappa_i
        = \frac{\lambda_{\min}}
               {\lambda_1 + \lambda_2 + \lambda_3}\in \left[0,\frac{1}{3} \right]. 
    \label{eq:curvature}
\end{equation}

Small values of $\kappa_i$ indicate that the neighbours of ${\mathbf{p}}_i$ lie approximately on a plane (low curvature), whereas large values characterise edges, ridges, or scattered noise. Points with excessive curvature are removed from the candidate ground set:

\begin{equation}
    \mathcal{P}_{\text{ground}}^{\kappa}
        = \left\{
            \mathbf{p}_i \in \mathcal{P}_{\text{ground}}^{\rho}
            \;\middle|\;
            \kappa_i \leq \kappa_{\max}
          \right\}.
    \label{eq:curvaturefilter}
\end{equation}

The threshold $\kappa_{\max}$ controls the tolerance to local surface irregularities.
 
\subsubsection{Region Growing Algorithm} \label{subsubsec:rg_algorithm}
 
The region growing algorithm verifies the spatial connectivity of the candidate ground set $\mathcal{P}_{\text{ground}}^{\kappa}$. The physical rationale is that a real navigable surface constitutes a single contiguous region; isolated clusters of points that pass the local geometric checks but are spatially disconnected from the main ground patch most likely
correspond to sensor artefacts or small flat-topped debris.
 
The seed point $p_s\in \mathbb{R}^3$ is selected as the point in $\mathcal{P}_{\text{ground}}^{\kappa}$ closest to the robot origin in the $xy$ plane of \texttt{base\_link}, since the ground directly beneath the robot is the most reliable reference:

\begin{equation}
    p_s
        = \underset{\mathbf{p}_i \,\in\,
                    \mathcal{P}_{\text{ground}}^{\kappa}}{\arg\min}
          \sqrt{x_i^2 + y_i^2}.
    \label{eq:seed}
\end{equation}
 
Starting from $\mathcal{R} = \{p_s\}$, a candidate point ${\mathbf{p}}_j \in \mathcal{P}_{\text{ground}}^{\kappa} \setminus \mathcal{R}$ is added to the region if it simultaneously satisfies three conditions: (i) its signed distance to the globally fitted plane does not exceed the ground tolerance $h_{\text{ground}}$ already defined in Eq.~\eqref{eq:classification}; (ii) its local normal is consistent with $\hat{\mathbf{n}}$ at level $\rho_{\min}$ (Eq.~\eqref{eq:normalfilter}); and (iii) it is spatially adjacent to at least one point already in $\mathcal{R}$ within a growing radius $r_g\in \mathbb{R}^+$:

\begin{equation}
    \mathbf{p}_j \text{ is added to } \mathcal{R}
    \iff
    \begin{cases}
        \left|\delta_j\right| \leq h_{\text{ground}}, \text{and} \\[4pt]
        \rho_j \geq \rho_{\min}, \text{and} \\[4pt]
        \exists\;\mathbf{p}_k \in \mathcal{R}:
            \left\|\mathbf{p}_j - \mathbf{p}_k\right\|
            \leq r_g
    \end{cases}
    \label{eq:growingcriterion}
\end{equation}
where $\delta_j = \hat{\mathbf{n}}^\top \mathbf{p}_j + \hat{d}$ is the signed distance of $\mathbf{p}_j$ to the fitted plane (Eq.~\eqref{eq:signeddist}), and $r_g$ is the spatial adjacency radius.

\subsubsection{Verified Ground and Obstacle Outputs} \label{subsubsec:rg_outputs}
 
The verified ground set $\mathcal{P}_{\text{ground}}^{\text{RG}}$ is defined as the connected region produced by the growing algorithm:

\begin{equation}
    \mathcal{P}_{\text{ground}}^{\text{RG}}
        \;=\; \mathcal{R}.
    \label{eq:rg_ground}
\end{equation}

\begin{figure}[!ht]
    \centering
    \subfloat[Greenhouse floor segmentation using the base algorithm \label{fig:Green1}]{
    \includegraphics[width=0.65\linewidth]{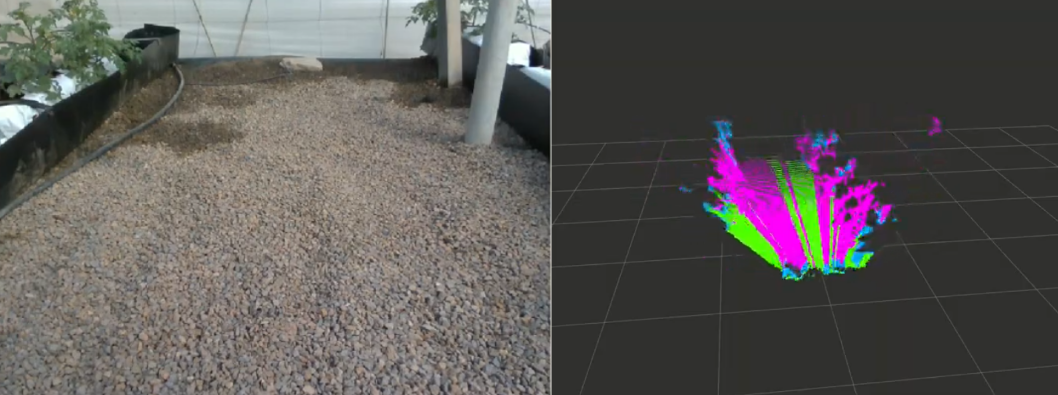}
    }
    \\
    \subfloat[GreenSeg greenhouse floor segmentation \label{fig:Green2}]{
        \includegraphics[width=0.65\linewidth]{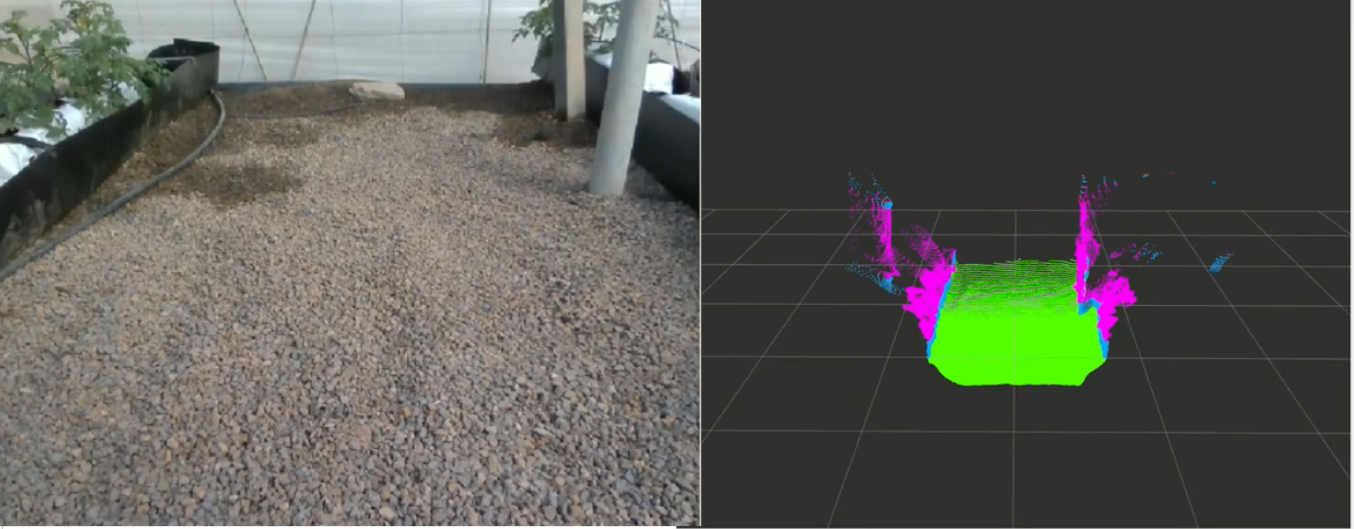}
        }
    \caption{Comparison of algorithms}
    \label{fig:Green}
\end{figure}

All points that were initially classified as ground by the base algorithm but did not survive the region growing validation (because they failed the normal consistency check, the curvature filter, or the spatial connectivity criterion) are conservatively promoted to the \textit{obstacle} or \textit{noise} class (depending $\kappa_i$ value). The verified obstacle cloud is therefore updated in the ROS 2 topic:

\begin{equation}
    \mathcal{P}_{\text{obs}}^{\text{RG}}
        \;=\; \mathcal{P}_{\text{obs}}
              \;\cup\;
              \left(
                  \mathcal{P}_{\text{ground}}^{(0)}
                  \setminus
                  \mathcal{P}_{\text{ground}}^{\text{RG}}
              \right),
    \label{eq:rg_obstacles}
\end{equation}

where $\mathcal{P}_{\text{obs}}$ is the obstacle cloud produced by the base
algorithm (Eq.~\eqref{eq:obstacles}). This conservative assignment ensures that any point whose ground membership cannot be spatially confirmed is treated as a potential collision threat, which is the appropriate behavior for delicate agricultural environments where a false negative (missed obstacle) is more costly than a false positive (over-cautious path planning). The outputs $\mathcal{P}_{\text{ground}}^{\text{RG}}$ and $\mathcal{P}_{\text{obs}}^{\text{RG}}$ are published in two new ROS~2 \texttt{sensor\_msgs/PointCloud2/greenseg} topics. The first constitutes the complete semantically labeled cloud. The performance enhancement relative to the baseline algorithm is illustrated in Figure~\ref{fig:Green}, where Figure~\ref{fig:Green1} depicts the baseline representation in contrast to the proposed approach within the same experimental trial compared with the new approach shown in the Figure \ref{fig:Green2}. In both cases, green indicates the ground, purple indicates obstacles, and cyan indicates points representing structures that are higher than the robot (such as the greenhouse roof, lamps or overhead cables).

\subsection{Analysis of the most adverse conditions in the Mediterranean Greenhouse}

The performance of the optical system is intrinsically linked to the geometric orientation of the incident light, defined by the sun’s elevation and azimuth angles \cite{jafarbiglu2023impact}. Furthermore, in a greenhouse, this reflection is increased as the polypropylene in the roof sheeting deflects the light, scattering it and making it more diffuse. 

\subsubsection{Analysis of solar lighting in the greenhouse}

Whilst both determine the dynamic range of the scene, their physical impact on the camera lens elements follows distinct phenomenological trajectories:

\begin{itemize}
    \item \textit{Sun Elevation ($\gamma$)}: This parameter has the most direct and disruptive influence on the lens's internal optics. At low elevation angles (e.g., during dawn or dusk), light rays strike the lens at grazing angles, significantly increasing the probability of lens flare, ghosting, and stray light reflections within the internal lens barrel \cite{honkavaara2012influence}. Furthermore, elevation dictates the intensity of specular reflections on the polyethylene covers of the greenhouse, which often leads to sensor saturation in the depth perceptual pipeline.
    \item \textit{Solar Azimuth ($\alpha$)}: While azimuth does not typically cause the same level of internal optical aberration as elevation, it fundamentally governs the spatial distribution of shadows and the orientation of high-contrast boundaries. In the narrow aisles of a greenhouse, the azimuth angle determines whether a crop row is illuminated frontally or back-lit. Back-lighting (high relative azimuth) creates "halo" effects around leaves and structures, challenging the edge-detection capabilities of the segmentation algorithm and potentially distorting the perceived geometry of the ground plane \cite{honkavaara2012influence,jafarbiglu2023impact}.
\end{itemize}

\begin{figure}[!ht]
    \centering
    \includegraphics[width=0.75\linewidth]{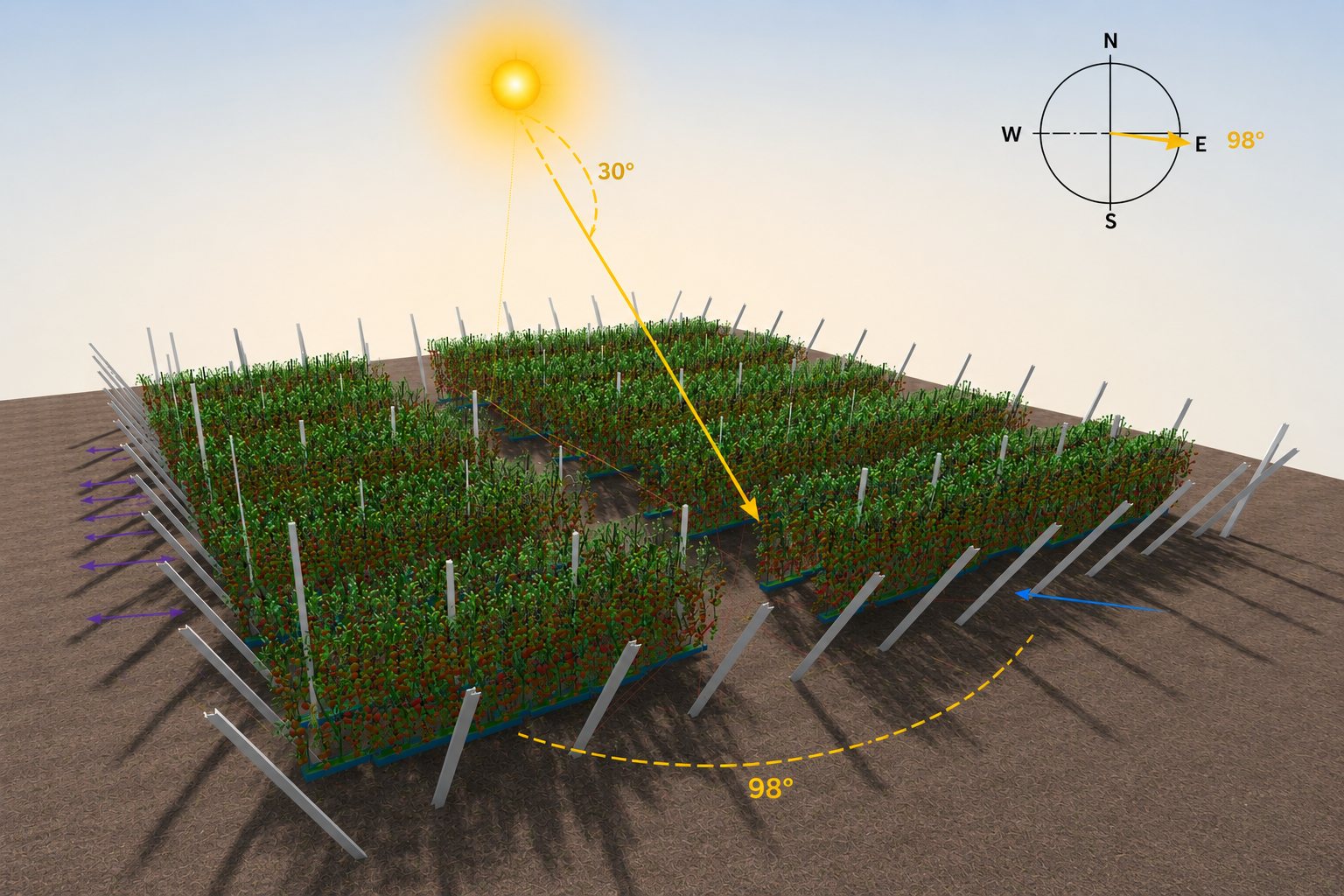}
    \caption{Example of solar elevation and azimuth in the greenhouse}
    \label{fig:Azimut}
\end{figure}

The irradiance phenomena and solar geometry affecting the perception system are inherently dictated by the geographical coordinates of the testbed (Figure \ref{fig:Azimut}).

For this study, the AgroConnect greenhouse is located in Almería, Spain ($36^{\circ}50'$~N, $2^{\circ}24'$~W). The experiments were carried out on 21th April 2026. To identify the critical operational windows, solar intensity profiles were retrieved from \textit{TheSkyLive} platform\footnote{TheSkyLive: \url{https://theskylive.com}}, while precise solar elevation and azimuth angles were computed using the \textit{SunEarthTools} ephemeris engine\footnote{SunEarthTools: \url{https://www.sunearthtools.com}}. Based on this spatial-temporal analysis, the performance of the proposed algorithm was evaluated under the following experimental presented in the Table \ref{tab:solar_scenarios}:

\begin{table}[!ht]
\centering
\caption{Solar geometry parameters at specific operational windows in 21/04/2026 in Almería, Spain.}
\label{tab:solar_scenarios}
\begin{tabular}{@{}lccc@{}}
\toprule
\textbf{Scenario} & \textbf{Time (UTC+2)} & \textbf{Sun Elevation ($\gamma$)} & \textbf{Solar Azimuth ($\alpha$)} \\ \midrule
1 & 08:00 h & 30$^{\circ}$ & 98$^{\circ}$ \\
2 & 10:00 h & 40$^{\circ}$ & 120$^{\circ}$ \\
3 & 12:00 h & 50$^{\circ}$ & 150$^{\circ}$ \\
4 & 14:00 h & 60$^{\circ}$ & 195$^{\circ}$ \\ \bottomrule
\end{tabular}
\end{table}

\subsubsection{Analysis of critical areas in the greenhouse}

The greenhouse environment presents unique structural challenges that directly impact the reliability of autonomous navigation systems. These critical areas are primarily defined by the interaction between the crop architecture and the incident solar radiation, which directly affects the performance of the proposed algorithm. First, the narrow crop aisles function as occlusion tunnels that depend on the solar azimuth and elevation discussed previously. Second, the greenhouse's plastic cladding induces non-uniform light scattering. Under high solar elevation conditions, specular saturation points occur on the polyethylene surfaces, adversely affecting depth sensors and creating "blind spots" within the point cloud. Furthermore, the areas adjacent to the lateral and frontal greenhouse openings are particularly critical due to the ingress of unfiltered direct sunlight, which induces optical artifacts (lens flare) in the camera sensors. Analyzing these zones is essential to validate the robustness of the proposed algorithm, as the operational performance is highly dependent on the specific area where the robot is operating and how these occlusions impact the system. The various scenarios typically encountered in a Mediterranean greenhouse, all of which serve as the basis for evaluating the algorithm's robustness, are: i) Moving along the central corridor; ii) Moving between rows; iii) Turning at the end of the corridor; and iv) Changing to the next corridor.

\subsection{Evaluation of the proposed approach} \label{subsec:metrics}

The performance of the proposed perception pipeline is assessed through a set of quantitative metrics computed over four field trials conducted in a Mediterranean greenhouse environment. Each test is compared with the original framework and the ground truth (GT); the methodology is detailed below.

\subsubsection{Ground-Truth identification}

In dynamic agricultural environments, static capture-based ground-truth generation is not representative of real operating conditions, since the robot navigates continuously over uneven soil, exposed roots, and scattered debris. Instead, a fully automatic ground-truth generation procedure is proposed, exploiting the temporal consistency of the raw point cloud
stream to identify high-confidence labels without manual intervention \cite{rosu2020semi}.

A semantic label is considered reliable for point $\tilde{\mathbf{p}}_i$ at time $t$ if and only if the geometric evidence supporting that label remains consistent across a temporal window of $W \in \mathbb{R}^+$ consecutive frames centered at $t$. Transient or ambiguous classifications caused by specular reflections, motion blur, or sensor noise generate inconsistent labels across frames and are therefore automatically excluded from the ground-truth set. To enable point-to-point correspondence across frames despite the robot's motion, the 3D space is discretized into a voxel grid with a resolution defined in the robot's \texttt{base\_link} reference frame. Each voxel accumulates the labels assigned by the algorithm to all points falling within it during the temporal window $[t - W/2,\; t + W/2]$. Voxels labeled as \textit{undefined} are excluded from all metric computations. For a window size of $W=10$ frames, a label is only accepted as a ground truth if a point has been labelled as a ground truth in at least 90\% of the total $W$ frames ($\psi_{min} = 0.9$) within the time window, where $\psi_{min}$ is the threshold value that determines when a point belongs to the ground truth.

\subsubsection{Evaluation metrics}

For each semantic class $k \in \mathcal{K} =  \{\textit{ground},\, \textit{obstacle},\, \textit{above},\, \textit{noise}\}$, the standard confusion-matrix quantities are defined over $\mathcal{D}_{\text{GT}}$ as:

\begin{itemize}
    \item $TP_k$ (true positives: points identified as correct in $k$ that are correct in the GT).
    \item $FP_k$ (false positives: points identified as correct in $k$ that are incorrect in the GT).
    \item $FN_k$ (false negatives: points identified as correct in $k$ but belonging to another class in GT).
    \item $TN_k$ (true negatives: points of another class correctly predicted as such).
\end{itemize}

To assess the accuracy $P_k \in \mathbb{R}^+$, the fraction of predicted positives for class $k$ that are correct is calculated:

\begin{equation}
    P_k = \frac{TP_k}{TP_k + FP_k}.
    \label{eq:precision}
\end{equation}

Furthermore, the recall for class $k$, denoted as $R_k \in [0, 1]$, quantifies the sensitivity of the model by representing the proportion of ground truth instances that are correctly identified:

\begin{equation}
    R_k = \frac{TP_k}{TP_k + FN_k}.
    \label{eq:recall}
\end{equation}

The harmonic mean of precision and recall, $F_{1,k}\in \mathbb{R}^+$, providing a single balanced measure, is given by:
\begin{equation}
    F_{1,k} = 2\cdot\frac{P_k \cdot R_k}{P_k + R_k}
        = \frac{2\,TP_k}{2\,TP_k + FP_k + FN_k}.
    \label{eq:f1}
\end{equation}

The standard overlap metric for semantic segmentation is the Intersection over Union (IoU) $\text{IoU}_k\in \mathbb{R}^+$, measuring the ratio of correctly classified points to the total points involved for class $k$:

\begin{equation}
    \text{IoU}_k = \frac{TP_k}{TP_k + FP_k + FN_k}.
    \label{eq:iou}
\end{equation}

\section{Results and Discussion} \label{sec: result}

This section describes the simulations and experiments designed to validate the proposed approach, as well as the results obtained.

\subsection{Performance Under Critical Zones and Solar Conditions}

\begin{figure}[!ht]
    \centering
    \includegraphics[width=0.7\linewidth]{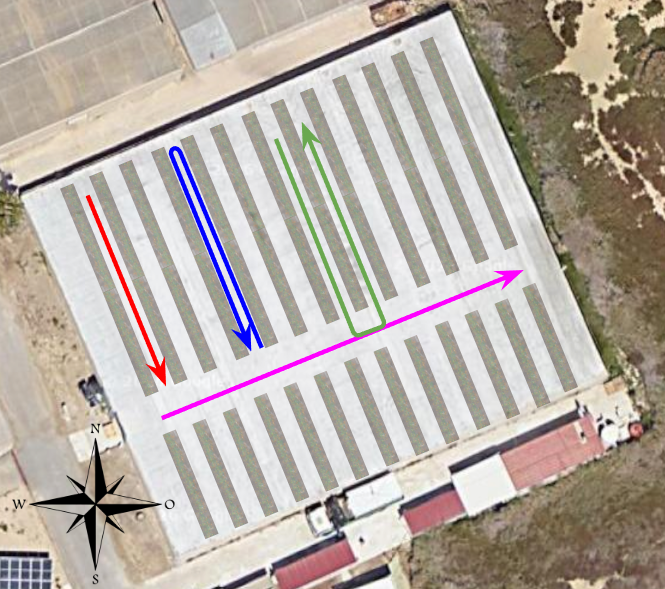}
    \caption{Tests paths. The red line represents the path moving through the central corridor; the red line represents the path moving through the rows; the blue line represents the path turning at the end of the corridor; the green line represents the path changing to the next corridor}
    \label{fig:Path}
\end{figure}

No dynamic obstacles were present during this trial. The terrain consisted of dry, homogeneous concrete, and the robot navigated at a constant speed of 1 m/s, while the ROS 2 perception node processed data at 10 Hz. Figures \ref{fig:Path} and \ref{fig:1_} show the different paths followed by the robot and the visual results for the solar scenarios considered in Table \ref{tab:solar_scenarios}, respectively. The different tracks for the robot are:

\begin{itemize}
    \item \textit{Moving along the central corridor}: The robot navigates through the central concrete aisle, as indicated by the pink path in Figure \ref{fig:Path}. These trials were conducted at 08:00, 10:00, 12:00, and 14:00 h to assess the impact of the solar elevation and azimuth parameters defined in Table \ref{tab:solar_scenarios}.
    
    \item \textit{Moving between rows}: The robot traverses the tomato crop rows, following the red path shown in Figure \ref{fig:Path}. As in the previous case, all solar scenarios from Table \ref{tab:solar_scenarios} were evaluated.
    
    \item \textit{Turning at the end of the aisle}: The robot navigates through the crop rows and executes a turn at the end of the corridors to go back along the same corridor, represented by the blue path in Figure \ref{fig:Path}. All solar scenarios from Table \ref{tab:solar_scenarios} were tested.

    \item \textit{Changing to the adjacent corridor}: The robot moves across the crop rows and performs a transition to the next aisle, indicated by the green path in Figure \ref{fig:Path}. Similarly, all lighting conditions defined in Table \ref{tab:solar_scenarios} were evaluated.
\end{itemize}

The algorithm processes the depth information stream from the \texttt{/camera/\\camera/depth} topic, alongside the intrinsic and extrinsic parameters provided by the \texttt{/camera/camera/depth/camera\_info} topic. The intrinsic parameters utilized were those calibrated by the camera manufacturer, whereas the extrinsic parameters defining the spatial transformation relative to the robot's \texttt{base\_link} were manually configured at $(0.4, 0, 0.5)$~m. The specific configuration parameters used for the segmentation process are detailed in Table~\ref{tab:greenseg_parameters}.
\begin{table}[!ht]
\centering
\caption{GreenSeg algorithm parameters}
\label{tab:greenseg_parameters}
\renewcommand{\arraystretch}{1.2} % Añade un poco de espacio vertical para mejorar la lectura
\begin{tabularx}{\textwidth}{@{} l c X @{}}
\toprule
\textbf{Parameters} & \textbf{Value} & \textbf{Description} \\ \midrule
\texttt{max\_surface\_height} ($h_{\text{ground}}$) & 0.12 m & Maximum perpendicular tolerance for ground plane \\
\texttt{max\_incline} & $30.0^{\circ}$ & Maximum permissible inclination of the terrain \\
\texttt{robot\_height} ($h_{\text{robot}}$) & 0.5 m & Physical height of the AGRICOBIOT I platform \\ 
\texttt{n\_neighbors} & 30 & Minimum points for local normal estimation \\
\texttt{r\_neighbors} ($r$) & 0.05 m & Neighborhood search radius for PCA \\
\texttt{rho\_min} ($\rho_{\text{min}}$) & 0.90 & Minimum cosine similarity for normal consistency \\
\texttt{kappa\_max} ($\kappa_{\text{max}}$) & 0.05 & Maximum surface curvature threshold \\
\texttt{r\_growing} ($r_g$) & 0.05 m & Spatial adjacency radius for Region Growing \\ 
\texttt{max\_distance\_filtered} & 3.0 m & Maximum effective range for ground segmentation \\ 
\texttt{min\_distance\_filtered} & 0.3 m & Minimum depth cutoff to avoid self-occlusion \\ \bottomrule
\end{tabularx}
\end{table}

The results of the trials are shown in the following.

\subsubsection{Moving along the central corridor}

\begin{figure}[!ht]
    \centering
    \subfloat[Central corridor test at 8:00 h - 04/21/2026\label{fig:1_1}]{
    \includegraphics[width=0.45\linewidth]{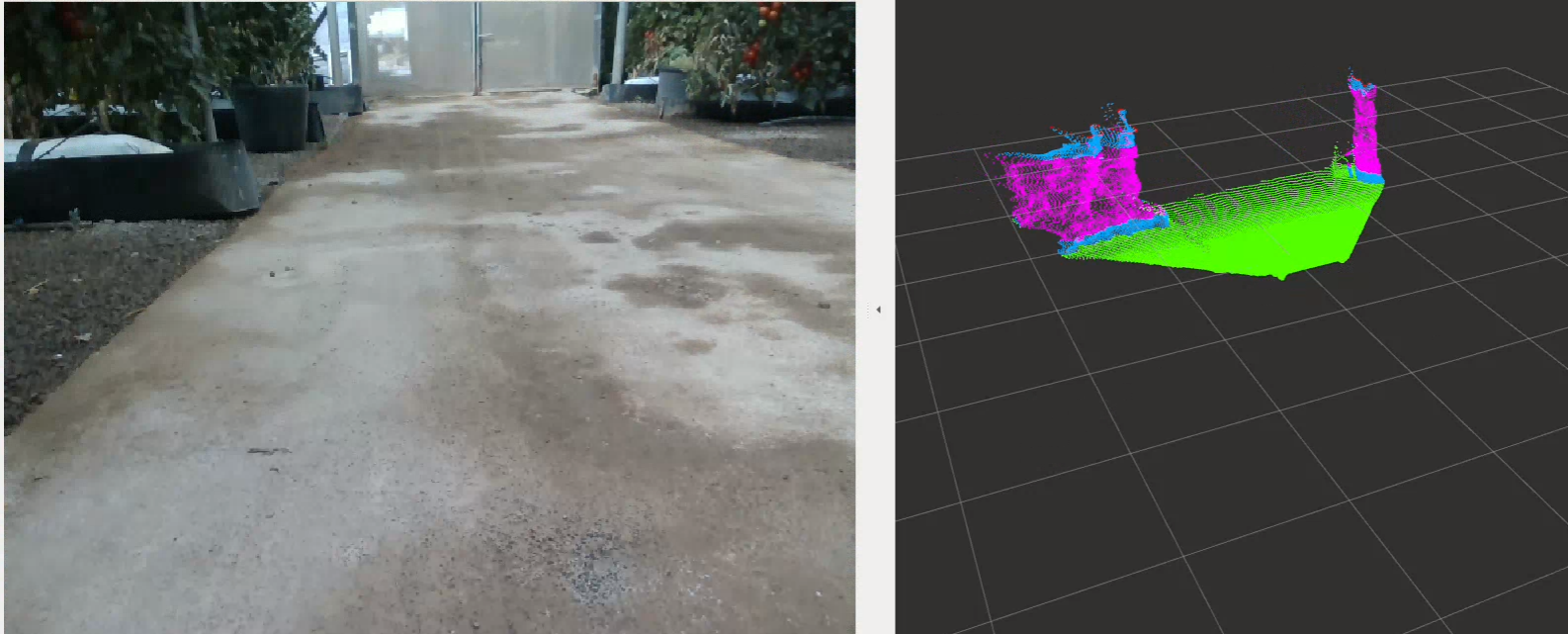}
    }
    \subfloat[Central corridor test at 10:00 h - 04/21/2026 \label{fig:1_2}]{
        \includegraphics[width=0.45\linewidth]{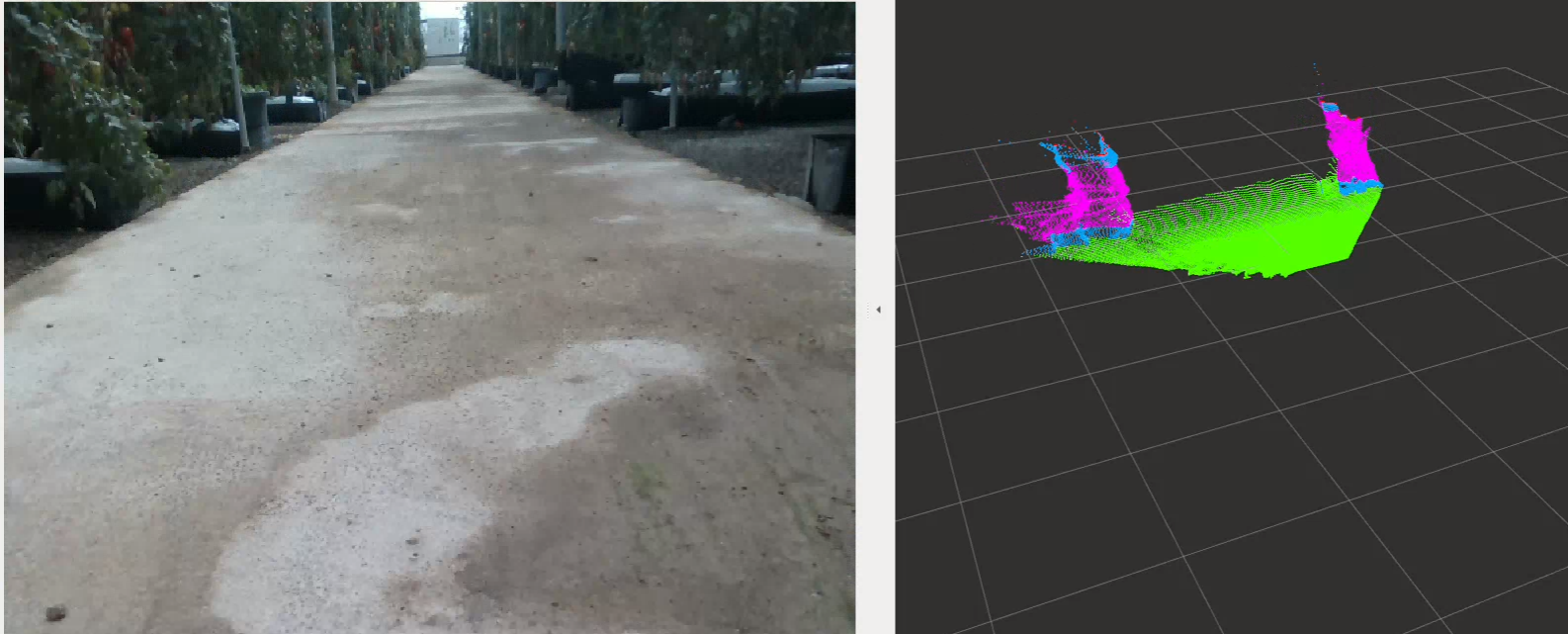}
        }
    \\
    \subfloat[Central corridor test at 12:00 h - 04/21/2026 \label{fig:1_3}]{
    \includegraphics[width=0.45\linewidth]{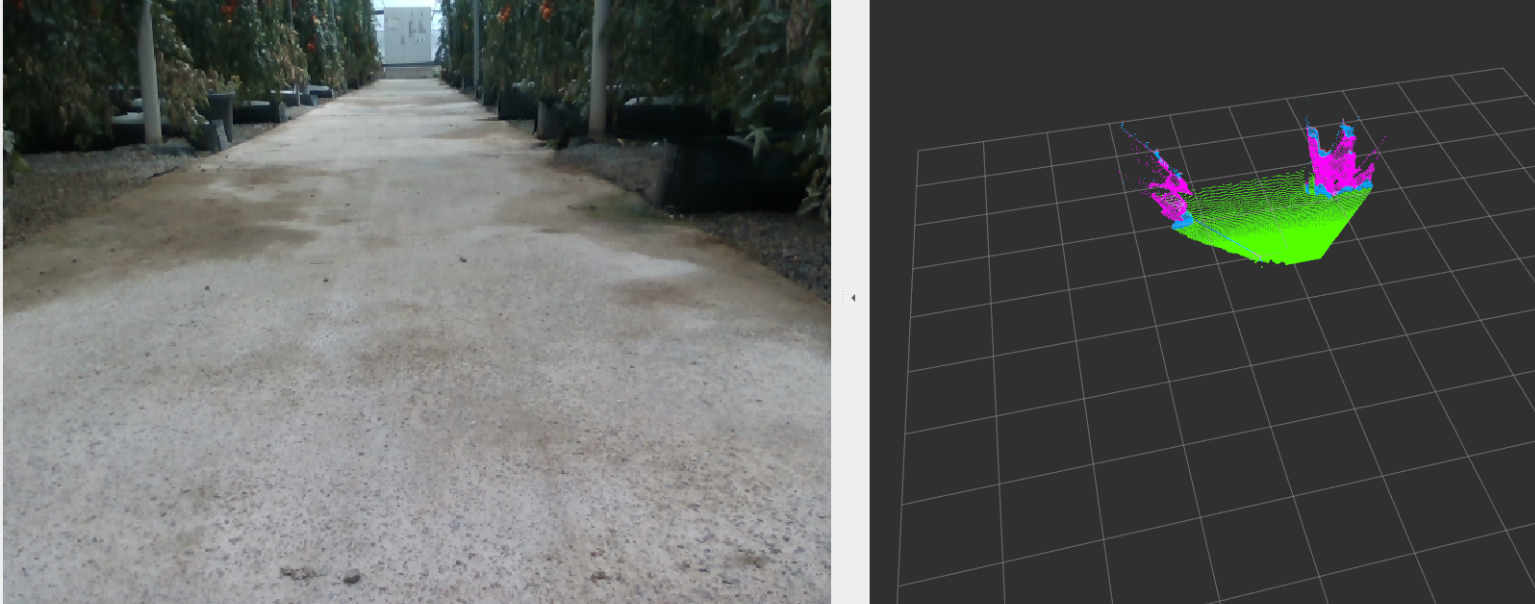}
    }
    \subfloat[Central corridor test at 14:00 h - 04/21/2026 \label{fig:1_4}]{
        \includegraphics[width=0.45\linewidth]{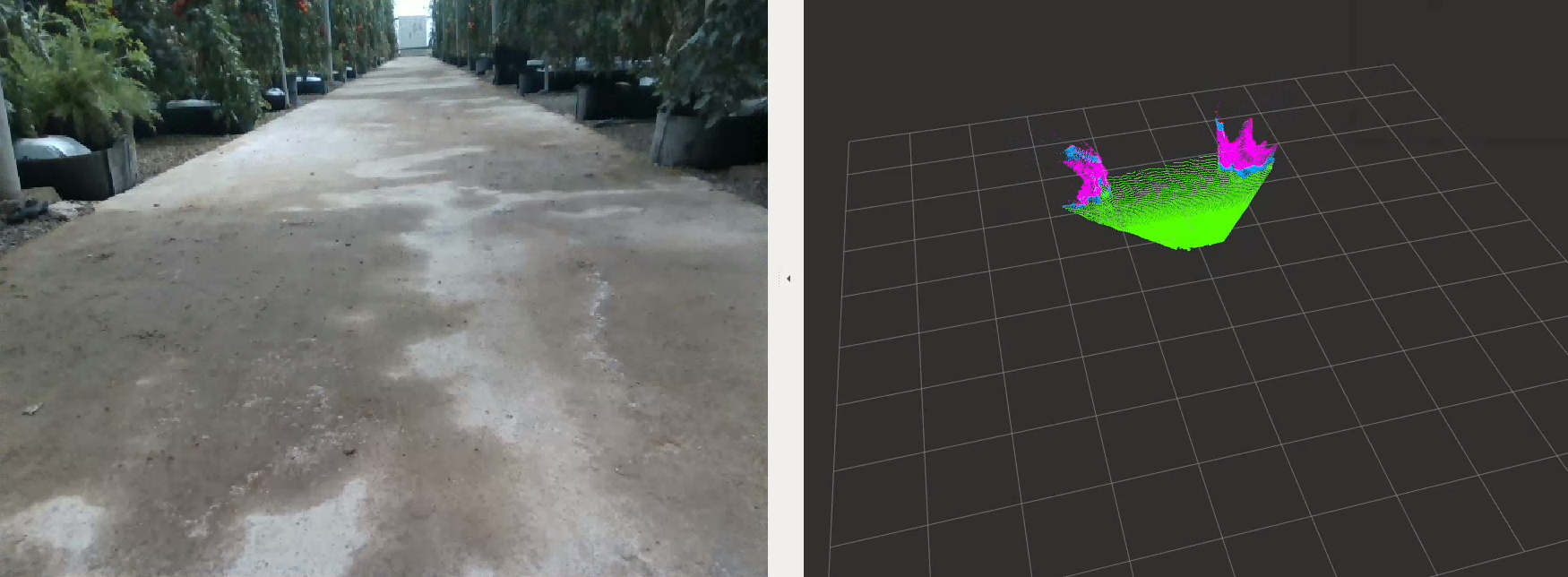}
        }
    \caption{Central corridor test at four solar conditions. Green points indicates the ground, purple indicates obstacles, and cyan indicates points are higher than the robot}
    \label{fig:1_}
\end{figure}

No dynamic obstacles were present, the terrain was dry and homogeneous, and the robot navigated at constant speed of 1 m/s. The ROS 2 node publishes the images at a rate of 10 Hz. The results at different times from this test are shown in the Figure \ref{fig:1_}.

As observed, the central corridor exhibits the most uniform terrain, composed of concrete to facilitate heavy machinery transit. Consequently, the perception framework demonstrates peak robustness in this area. The lateral aisles feature tomato plants in their 20th week of growth (reaching approximately a height of 2 meters) and robot is moving at constant speed of 0.5 m/s. The diurnal experimental analysis is detailed below:

\begin{itemize}
    \item \textit{Scenario 1 (08:00 h) - Figure \ref{fig:1_1}}: Low solar elevation ($\gamma = 30^{\circ}$) combined with the dense crop canopy prevents direct irradiance on the lens. GreenSeg correctly identifies the navigable plane, effectively filtering minor multipath noise near the corridor edges.
    \item \textit{Scenario 2 (10:00 h) - Figure \ref{fig:1_2}}: Although localized glare begins to emerge in the RGB stream, the geometric consistency check ($\rho_{min}$) prevents these optical artifacts from being misclassified as ground-level obstacles.
    \item \textit{Scenario 3 (12:00 h) - Figure \ref{fig:1_3}}: Despite increased solar irradiance amplified by the polyethylene cladding, the proposed second-layer validation maintains a stable IoU, significantly outperforming the baseline which struggles with floor-level specular reflections.
    \item \textit{Scenario 4 (14:00 h) - Figure \ref{fig:1_4}}: At peak solar intensity ($\gamma = 60^{\circ}$), the strategic placement of the camera beneath the AGRICOBIOT I transport platform provides passive optical shielding. This physical protection, combined with the Region Growing connectivity constraint, ensures that isolated 'ghost points' are successfully rejected.
\end{itemize}

Table~\ref{tab:trial1} reports all metrics defined in Section~\ref{subsec:metrics}. These results correspond to the average value of the experiment, with each data point classified within its respective class. This process is repeated for each subsequent experiment.

\begin{table}[!ht]
\centering
\caption{Results between baseline and GreenSeg for Test 1: Moving along the central corridor}
\label{tab:trial1}
\renewcommand{\arraystretch}{1.2}
\setlength{\tabcolsep}{2pt} % Reduce ligeramente el espacio fijo intercolumnar para dar margen de maniobra

% Definimos localmente la columna 'C' (requiere el paquete array, cargado automáticamente por tabularx)
\newcolumntype{C}{>{\centering\arraybackslash}X}

\begin{tabularx}{\textwidth}{@{} l *{8}{C} @{}}
\toprule
\multirow{2}{*}{\textbf{Class} $k$} & \multicolumn{2}{c}{\textbf{$P_k$}} & \multicolumn{2}{c}{\textbf{$R_k$}} & \multicolumn{2}{c}{\textbf{$F_{1,k}$}} & \multicolumn{2}{c}{\textbf{$\text{IoU}_k$}} \\
\cmidrule(lr){2-3} \cmidrule(lr){4-5} \cmidrule(lr){6-7} \cmidrule(lr){8-9}
& Base & Ours & Base & Ours & Base & Ours & Base & Ours \\ 
\midrule
\textit{ground}   & 0.956 & \textbf{0.970} & 0.948 & \textbf{0.960} & 0.958 & \textbf{0.965} & 0.931 & \textbf{0.933} \\
\textit{obstacle} & 0.891 & \textbf{0.910} & 0.915 & \textbf{0.970} & 0.915 & \textbf{0.939} & 0.865 & \textbf{0.885} \\
\textit{above}    & 0.935 & \textbf{0.940} & 0.926 & \textbf{0.930} & 0.931 & \textbf{0.935} & 0.875 & \textbf{0.878} \\
\textit{noise}    & 0.856 & \textbf{0.880} & 0.815 & \textbf{0.820} & 0.822 & \textbf{0.849} & 0.715 & \textbf{0.737} \\
\midrule
\textbf{Mean}     & 0.910 & \textbf{0.925} & 0.901 & \textbf{0.920} & 0.907 & \textbf{0.922} & 0.847 & \textbf{0.858} \\
\midrule
\textbf{Improvement} & \multicolumn{2}{c}{+1.65\%} & \multicolumn{2}{c}{+2.11\%} & \multicolumn{2}{c}{+1.65\%} & \multicolumn{2}{c}{+1.30\%} \\
\bottomrule
\end{tabularx}
\end{table}

The central corridor represents the most structured and least challenging scenario, characterized by flat terrain and minimal overhanging obstacles. Consequently, the baseline algorithm performs optimally, establishing a highly competitive benchmark. Nevertheless, GreenSeg still provides a measurable improvement, enhancing the mean IoU by +1.30\% and the overall obstacle recall by +2.11\%. This demonstrates that even in nominal conditions where geometric heuristics are generally sufficient, the proposed normal consistency and region-growing checks refine the segmentation boundaries without degrading the baseline's inherent stability.

\subsubsection{Moving between plants}

\begin{figure}[!ht]
    \centering
    \subfloat[Crop-row traversal at 08:00~h -- 04/21/2026\label{fig:2_1}]{
    \includegraphics[width=0.45\linewidth]{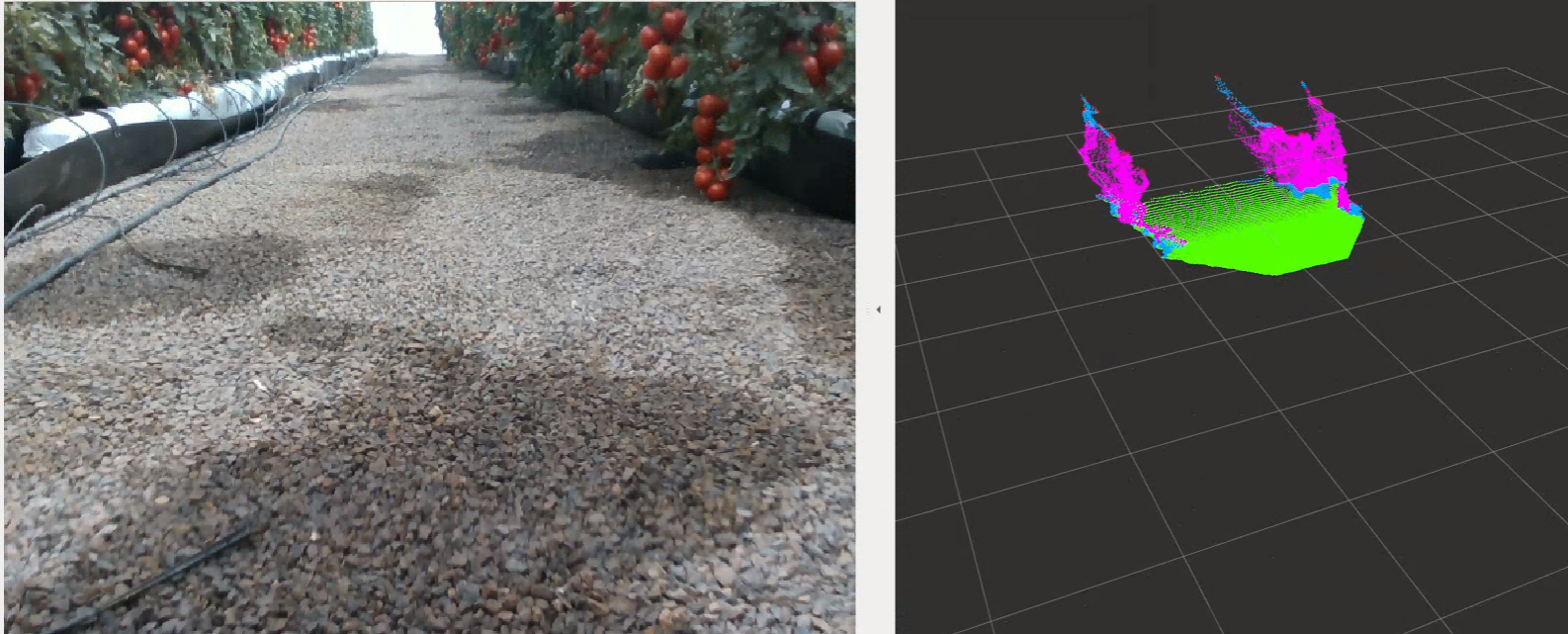}
    }
    \subfloat[Crop-row traversal at 10:00~h -- 04/21/2026 \label{fig:2_2}]{
        \includegraphics[width=0.45\linewidth]{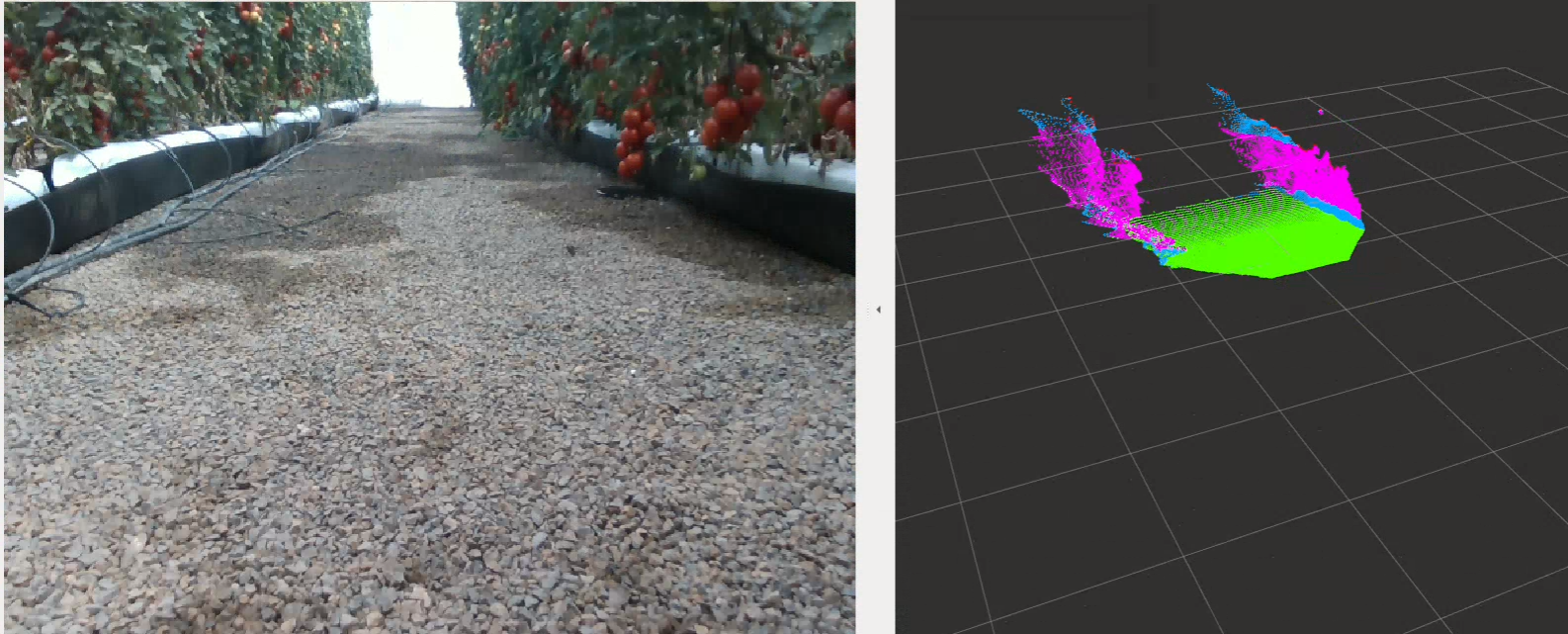}
        }
    \\
    \subfloat[Crop-row traversal at 12:00~h -- 04/21/2026 \label{fig:2_3}]{
    \includegraphics[width=0.45\linewidth]{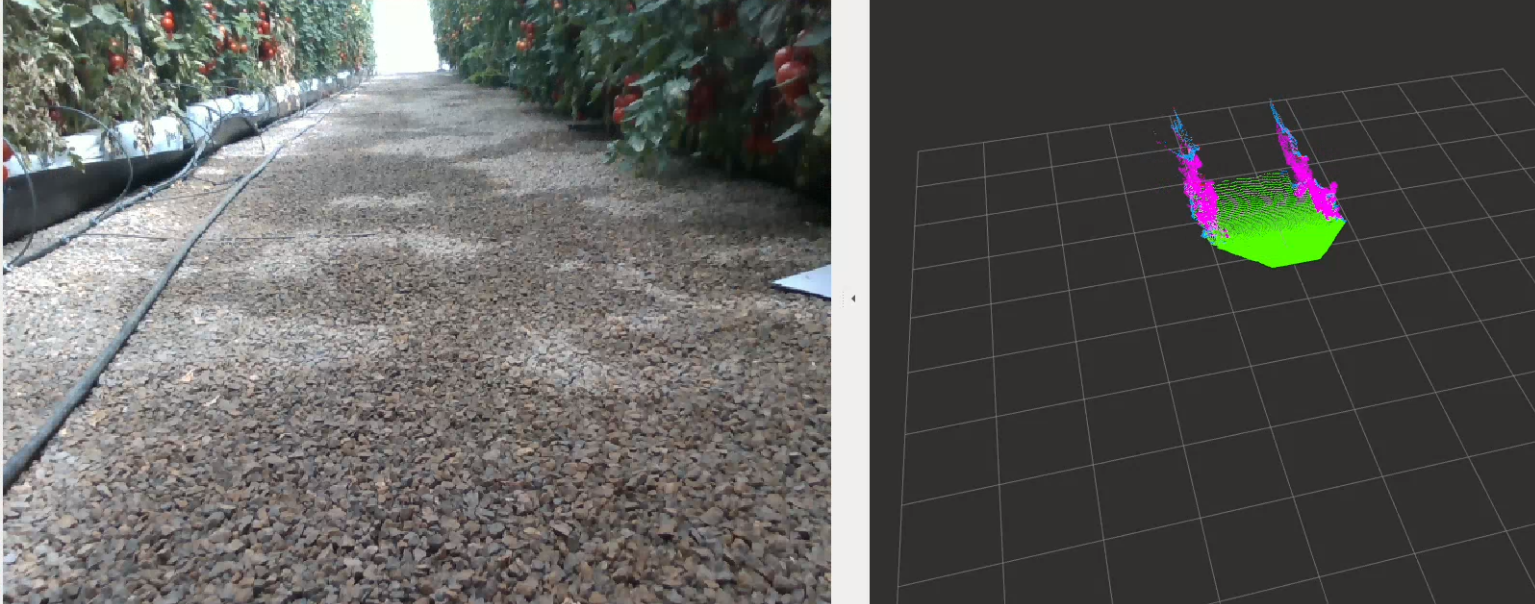}
    }
    \subfloat[Crop-row traversal at 14:00~h -- 04/21/2026 \label{fig:2_4}]{
        \includegraphics[width=0.45\linewidth]{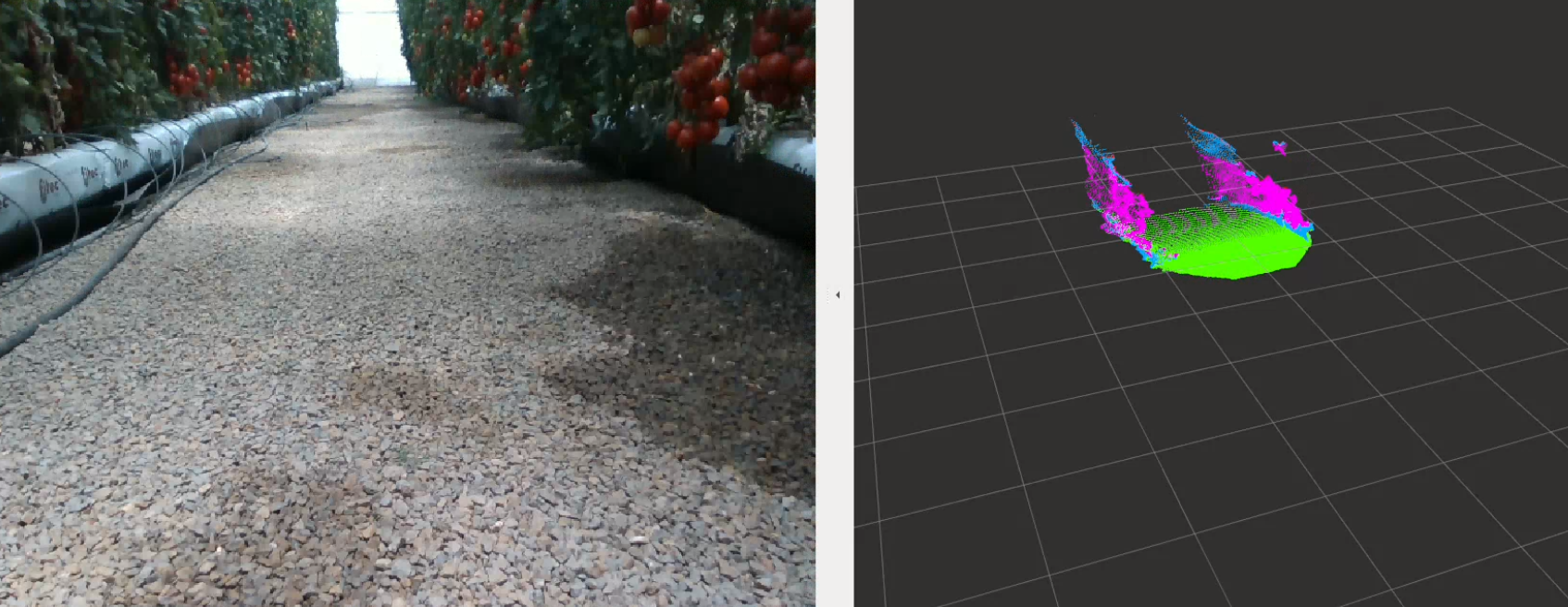}
        }
    \caption{Crop-row traversal test at four solar conditions. Green points indicates the ground, purple indicates obstacles, and cyan indicates points are higher than the robot}
    \label{fig:2_}
\end{figure}

This trajectory imposes greater perceptual complexity than the central corridor, as the ground transitions from smooth concrete to irregular gravel and tilled soil. Additionally, the navigable path is partially occluded by the lower canopy of the tomato plant. During this trial, the AGRICOBIOT I platform maintained a constant speed of 0.5 m/s to navigate the narrow 1 m width aisles. The visual results across different solar conditions are illustrated in Figure~\ref{fig:2_}.

In the case of the reference algorithm, the greater heterogeneity of the terrain on gravel results in localised variations in height that the reference algorithm often misclassifies as obstacles. However, the integration of the surface curvature filter ($\kappa_{max}$) into GreenSeg allows for a more flexible, yet geometrically consistent, identification of the ground. The daytime analysis is as follows:

\begin{itemize}
    \item \textit{Scenario 1 (08:00 h) - Figure \ref{fig:2_1}}: The low sun elevation and shadowing from the 2 m high plants provide an ideal contrast. GreenSeg successfully identifies the gravel plane despite the irregular texture.
    \item \textit{Scenario 2 (10:00 h) - Figure \ref{fig:2_2}}: Despite emerging glare, the spatial connectivity constraint of the Region Growing algorithm prevents isolated reflections on the plastic mulch from being segmented as navigable ground.
    \item \textit{Scenario 3 (12:00 h) - Figure \ref{fig:2_3}}: Peak light scattering through the polyethylene cover increases depth noise. GreenSeg compensates for this by using the local normal consistency check ($\rho_{min}$), filtering artefacts that do not align with the estimated ground plane.
    \item \textit{Scenario 4 (14:00 h) - Figure \ref{fig:2_4}}: The overhead sun ($\gamma=60^{\circ}$) creates harsh shadows between plants.
\end{itemize}

Quantitative results are summarized in Table~\ref{tab:test_ii}.
\begin{table}[!ht]
\centering
\caption{Results between baseline and GreenSeg for Test 2: Moving between crop rows}
\label{tab:test_ii}
\renewcommand{\arraystretch}{1.2}
\setlength{\tabcolsep}{2pt} % Reduce el espaciado fijo para prevenir desbordamientos de celda

% Definición local del tipo de columna flexible y centrada
\newcolumntype{C}{>{\centering\arraybackslash}X}

\begin{tabularx}{\textwidth}{@{} l *{8}{C} @{}}
\toprule
\multirow{2}{*}{\textbf{Class} $k$} & \multicolumn{2}{c}{\textbf{$P_k$}} & \multicolumn{2}{c}{\textbf{$R_k$}} & \multicolumn{2}{c}{\textbf{$F_{1,k}$}} & \multicolumn{2}{c}{\textbf{$\text{IoU}_k$}} \\
\cmidrule(lr){2-3} \cmidrule(lr){4-5} \cmidrule(lr){6-7} \cmidrule(lr){8-9}
& Base & Ours & Base & Ours & Base & Ours & Base & Ours \\ 
\midrule
\textit{ground}   & 0.885 & \textbf{0.935} & 0.870 & \textbf{0.925} & 0.877 & \textbf{0.930} & 0.781 & \textbf{0.869} \\
\textit{obstacle} & 0.815 & \textbf{0.885} & 0.830 & \textbf{0.935} & 0.822 & \textbf{0.909} & 0.698 & \textbf{0.833} \\
\textit{above}    & 0.860 & \textbf{0.910} & 0.855 & \textbf{0.895} & 0.857 & \textbf{0.902} & 0.750 & \textbf{0.821} \\
\textit{noise}    & 0.780 & \textbf{0.845} & 0.740 & \textbf{0.805} & 0.759 & \textbf{0.824} & 0.612 & \textbf{0.701} \\
\midrule
\textbf{Mean}     & 0.835 & \textbf{0.894} & 0.824 & \textbf{0.890} & 0.829 & \textbf{0.891} & 0.710 & \textbf{0.806} \\
\midrule
\textbf{Improvement} & \multicolumn{2}{c}{+7.06\%} & \multicolumn{2}{c}{+8.00\%} & \multicolumn{2}{c}{+7.47\%} & \multicolumn{2}{c}{+13.52\%} \\
\bottomrule
\end{tabularx}
\end{table}

In this environment, the baseline's column-wise scanning struggles with depth ambiguities caused by overhanging leaves, dynamic shadows, and irregular soil. By contrast, GreenSeg leverages region-growing connectivity to group leaf masses and continuous ground patches effectively. This architectural advantage is reflected in a substantial +13.52\% increase in mean IoU and an +8.00\% surge in overall Recall. Most importantly, the obstacle recall peaks at 0.935, proving that the proposed algorithm significantly reduces false negatives in dense vegetation, a critical factor for safe autonomous navigation.

\subsubsection{Turning at the end of the corridor}

\begin{figure}[!ht]
    \centering
    \subfloat[End-of-corridor turning at 08:00~h -- 04/21/2026\label{fig:3_1}]{
    \includegraphics[width=0.45\linewidth]{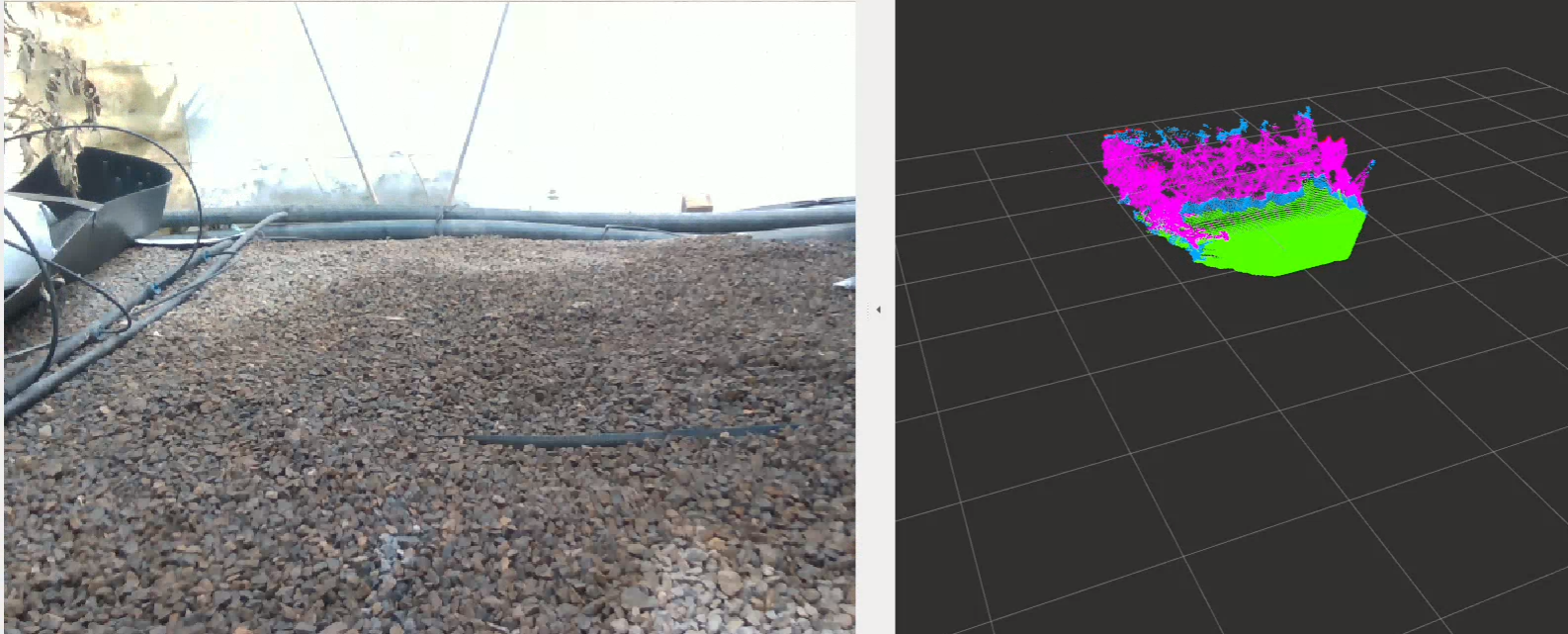}
    }
    \subfloat[End-of-corridor turning at 10:00~h -- 04/21/2026 \label{fig:3_2}]{
        \includegraphics[width=0.45\linewidth]{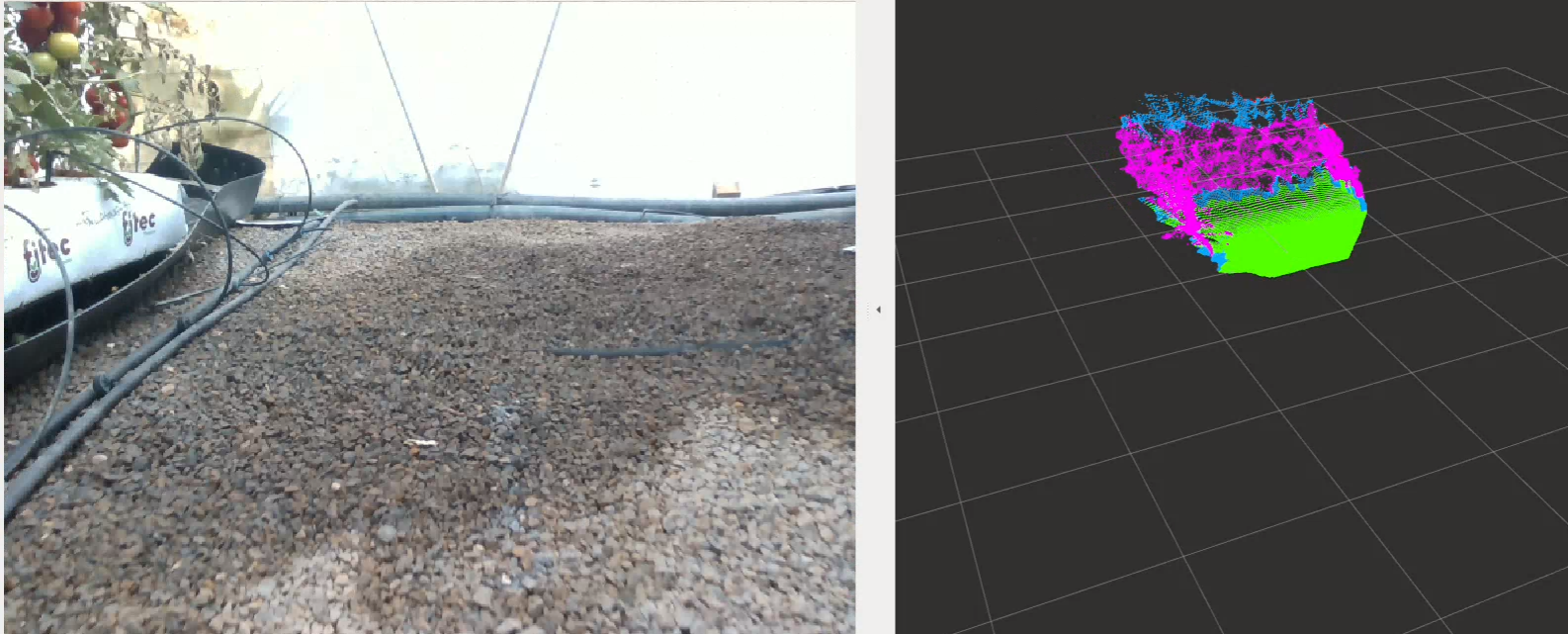}
        }
    \\
    \subfloat[End-of-corridor turning at 12:00~h -- 04/21/2026 \label{fig:3_3}]{
    \includegraphics[width=0.45\linewidth]{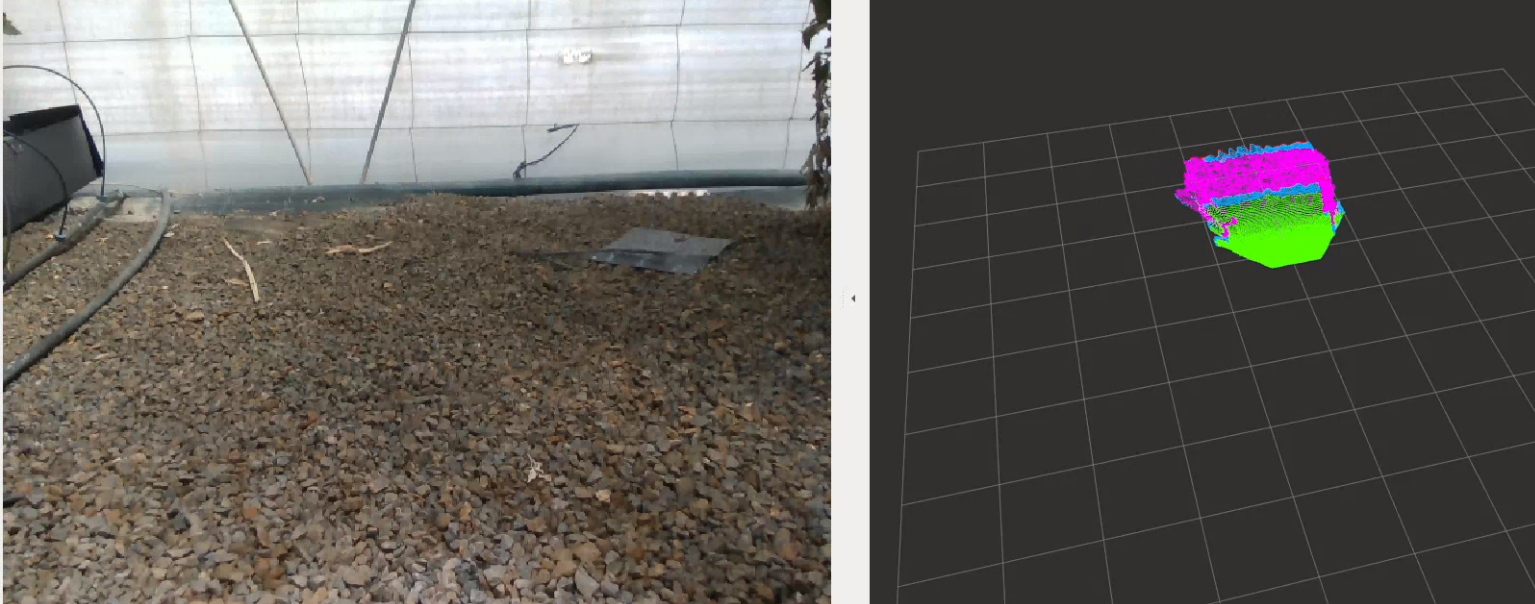}
    }
    \subfloat[End-of-corridor turning at 14:00~h -- 04/21/2026 \label{fig:3_4}]{
        \includegraphics[width=0.45\linewidth]{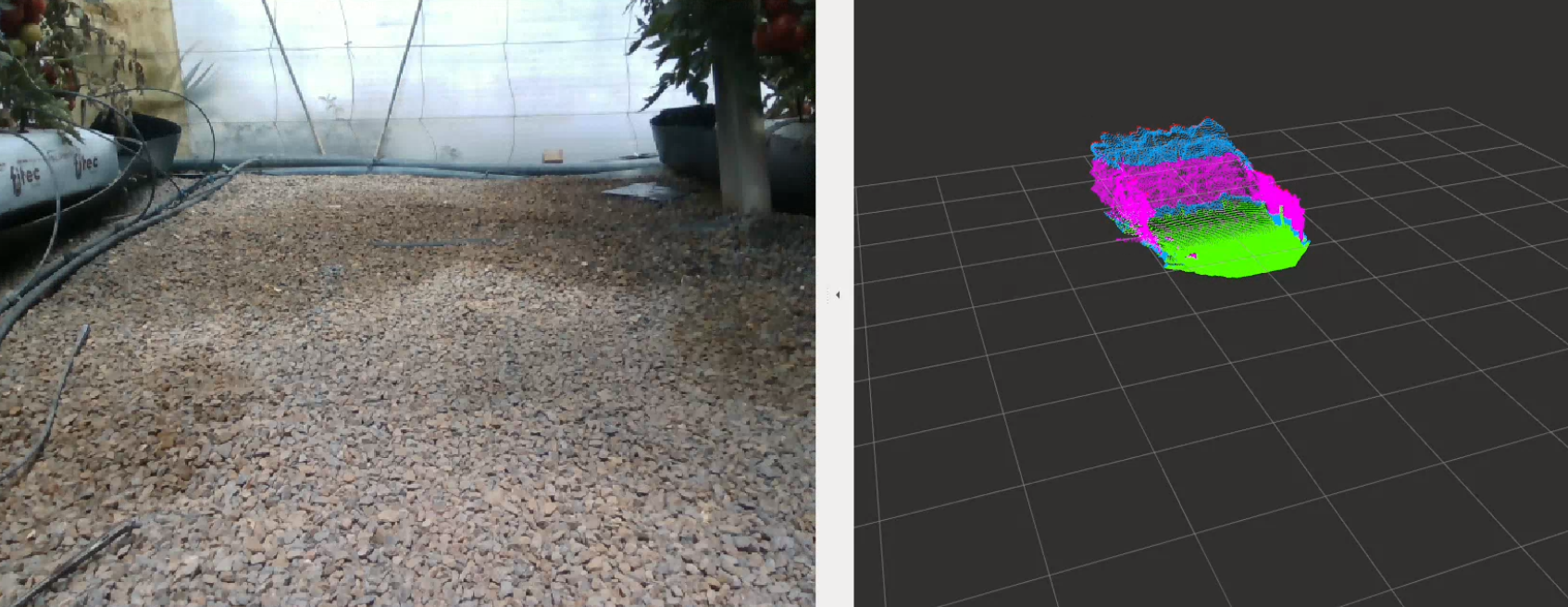}
        }
    \caption{End-of-corridor turning test at four solar conditions. Green points indicates the ground, purple indicates obstacles, and cyan indicates points are higher than the robot}
    \label{fig:3_}
\end{figure}

The end-of-aisle turning maneuver constitutes one of the most challenging scenarios for ground segmentation. During this phase, the robot reorients within a confined space while facing the frontal greenhouse openings, which admit unfiltered direct sunlight. This configuration frequently induces severe lens flare and abrupt depth discontinuities. Additionally, the rotational motion introduces transient noise that challenges temporal consistency. The visual results for this trial are shown in Figure~\ref{fig:3_}.

The experimental results for this manoeuvre are highly positive, as the proposed GreenSeg algorithm demonstrates superior capacity to maintain structural coherence compared to the baseline algorithm. By linking the ground plane to the robot's immediate vicinity via the seed point $p_s$, the system effectively overcomes the transition between shaded and overexposed areas. The diurnal analysis is as follows:

\begin{itemize}
    \item \textit{Scenario 1 (08:00 h) - Figure \ref{fig:3_1}}: Despite the low solar angle, the crop rows act as a natural blind. GreenSeg correctly identifies the transition to the turning area, maintaining high geometric stability with an $F_{1,ground}$.
    
    \item \textit{Scenario 2 (10:00 h) - Figure \ref{fig:3_2}}: As the robot faces the greenhouse opening, glare artifacts appear. The normal consistency score ($\rho_{min}$) effectively rejects these virtual points.
    
    \item \textit{Scenario 3 (12:00 h) - Figure \ref{fig:3_3}}: Under maximum irradiance, the proposed second-layer validation outperforms the baseline, maintaining the segmentation of the navigable surface even when the polyethylene cover amplifies the scattering.
    
    \item \textit{Scenario 4 (14:00 h) - Figure \ref{fig:3_4}}: At peak solar intensity, the Region Growing algorithm ensures spatial continuity.
    
\end{itemize}
    
Quantitative results reported in Table~\ref{tab:test_iii} confirm the effectiveness of the approach:

\begin{table}[!ht]
\centering
\caption{Results between baseline and GreenSeg for Test 3: Turning at the end of the corridor}
\label{tab:test_iii}
\renewcommand{\arraystretch}{1.2}
\setlength{\tabcolsep}{2pt} % Optimiza el espaciado interno de las celdas

% Definición del tipo de columna flexible centrada (asegúrate de que no esté duplicado en tu preámbulo)
\newcolumntype{C}{>{\centering\arraybackslash}X}

\begin{tabularx}{\textwidth}{@{} l *{8}{C} @{}}
\toprule
\multirow{2}{*}{\textbf{Class} $k$} & \multicolumn{2}{c}{\textbf{$P_k$}} & \multicolumn{2}{c}{\textbf{$R_k$}} & \multicolumn{2}{c}{\textbf{$F_{1,k}$}} & \multicolumn{2}{c}{\textbf{$\text{IoU}_k$}} \\
\cmidrule(lr){2-3} \cmidrule(lr){4-5} \cmidrule(lr){6-7} \cmidrule(lr){8-9}
& Base & Ours & Base & Ours & Base & Ours & Base & Ours \\ 
\midrule
\textit{ground}   & 0.810 & \textbf{0.890} & 0.805 & \textbf{0.885} & 0.807 & \textbf{0.887} & 0.677 & \textbf{0.796} \\
\textit{obstacle} & 0.740 & \textbf{0.840} & 0.765 & \textbf{0.890} & 0.752 & \textbf{0.864} & 0.602 & \textbf{0.760} \\
\textit{above}    & 0.795 & \textbf{0.855} & 0.780 & \textbf{0.845} & 0.787 & \textbf{0.849} & 0.649 & \textbf{0.738} \\
\textit{noise}    & 0.690 & \textbf{0.780} & 0.655 & \textbf{0.735} & 0.672 & \textbf{0.756} & 0.506 & \textbf{0.608} \\
\midrule
\textbf{Mean}     & 0.758 & \textbf{0.841} & 0.751 & \textbf{0.838} & 0.754 & \textbf{0.839} & 0.608 & \textbf{0.725} \\
\midrule
\textbf{Improvement} & \multicolumn{2}{c}{+10.94\%} & \multicolumn{2}{c}{+11.58\%} & \multicolumn{2}{c}{+11.27\%} & \multicolumn{2}{c}{+19.24\%} \\
\bottomrule
\end{tabularx}
\end{table}

During end-of-aisle turns, the camera's rotation introduces severe motion artifacts and disrupts the assumption of a continuous forward-facing slope, causing the baseline's column-wise logic to fail and fragment the navigable space. GreenSeg mitigates this through RANSAC-based global plane fitting, which maintains a mathematical reference of the ground regardless of the camera's yaw rate. This robustness yields the most dramatic performance gains in the study: a +11.58\% improvement in mean Recall and a remarkable +19.24\% boost in mean IoU, ensuring the local costmap remains coherent precisely when the robot is most susceptible to collisions.

\subsubsection{Changing to the next corridor}

\begin{figure}[!ht]
    \centering
    \subfloat[Changing to the next corridor at 08:00~h -- 04/21/2026\label{fig:4_1}]{
    \includegraphics[width=0.45\linewidth]{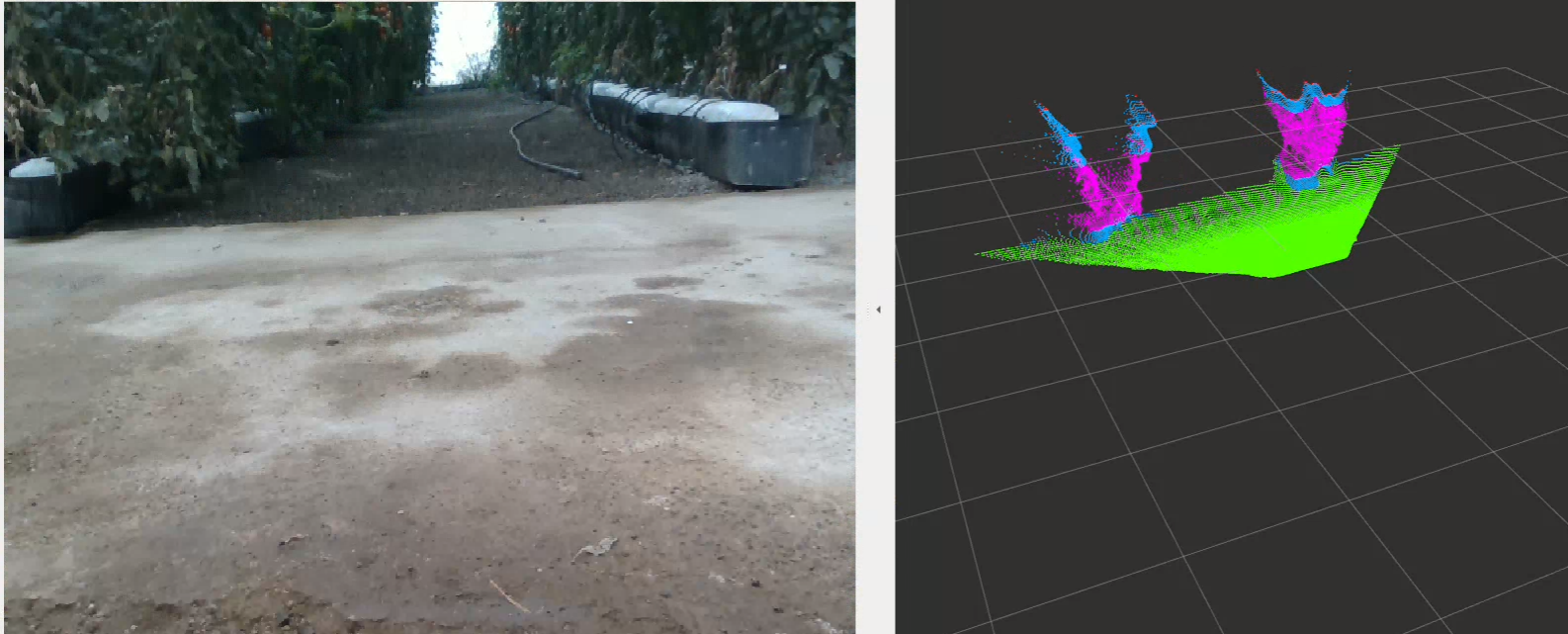}
    }
    \subfloat[Changing to the next corridor at 10:00~h -- 04/21/2026 \label{fig:4_2}]{
        \includegraphics[width=0.45\linewidth]{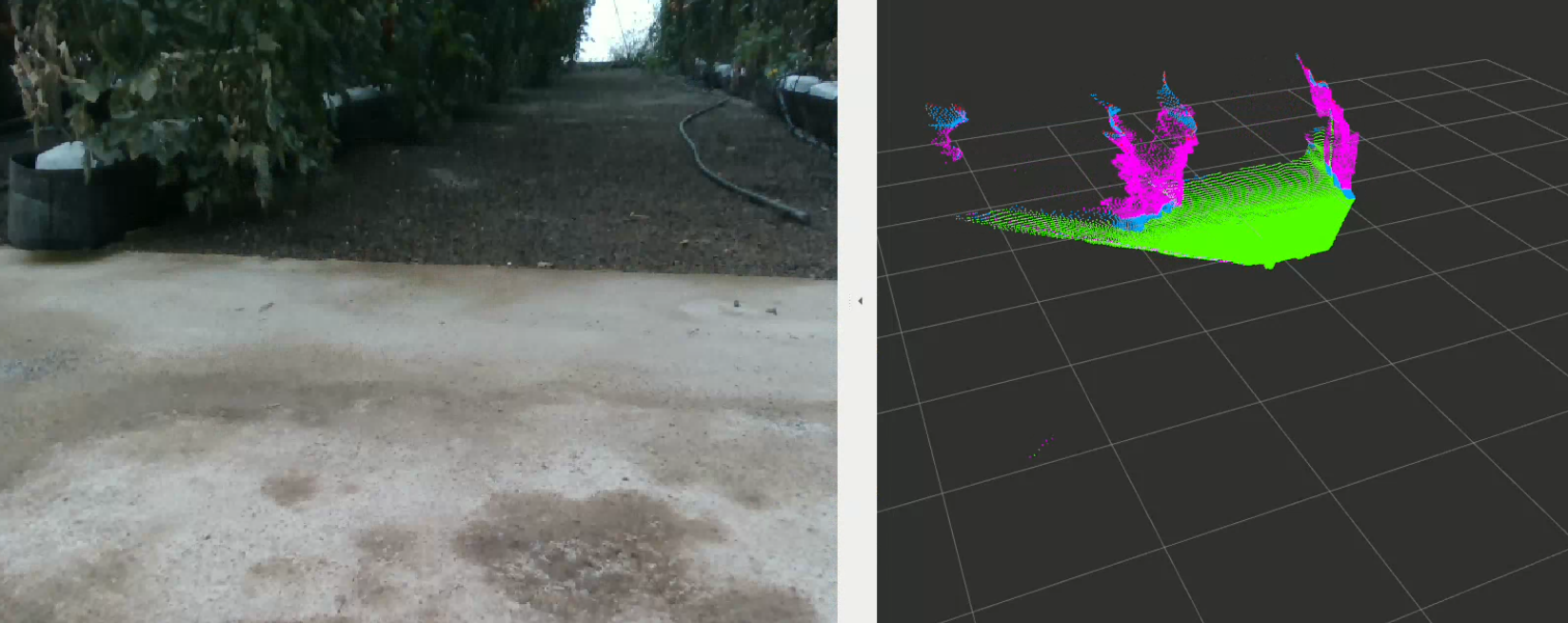}
        }
    \\
    \subfloat[Changing to the next corridor at 12:00~h -- 04/21/2026 \label{fig:4_3}]{
    \includegraphics[width=0.45\linewidth]{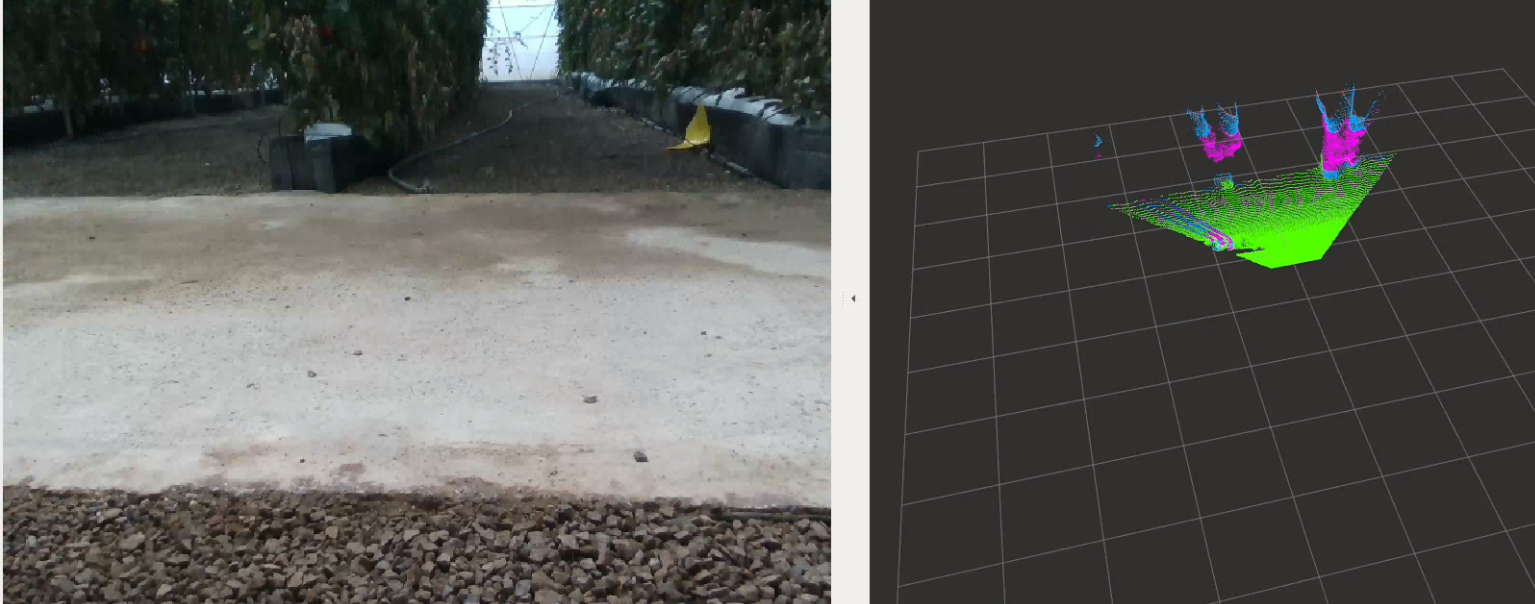}
        }
    \subfloat[Changing to the next corridor at 14:00~h -- 04/21/2026 \label{fig:4_4}]{
        \includegraphics[width=0.45\linewidth]{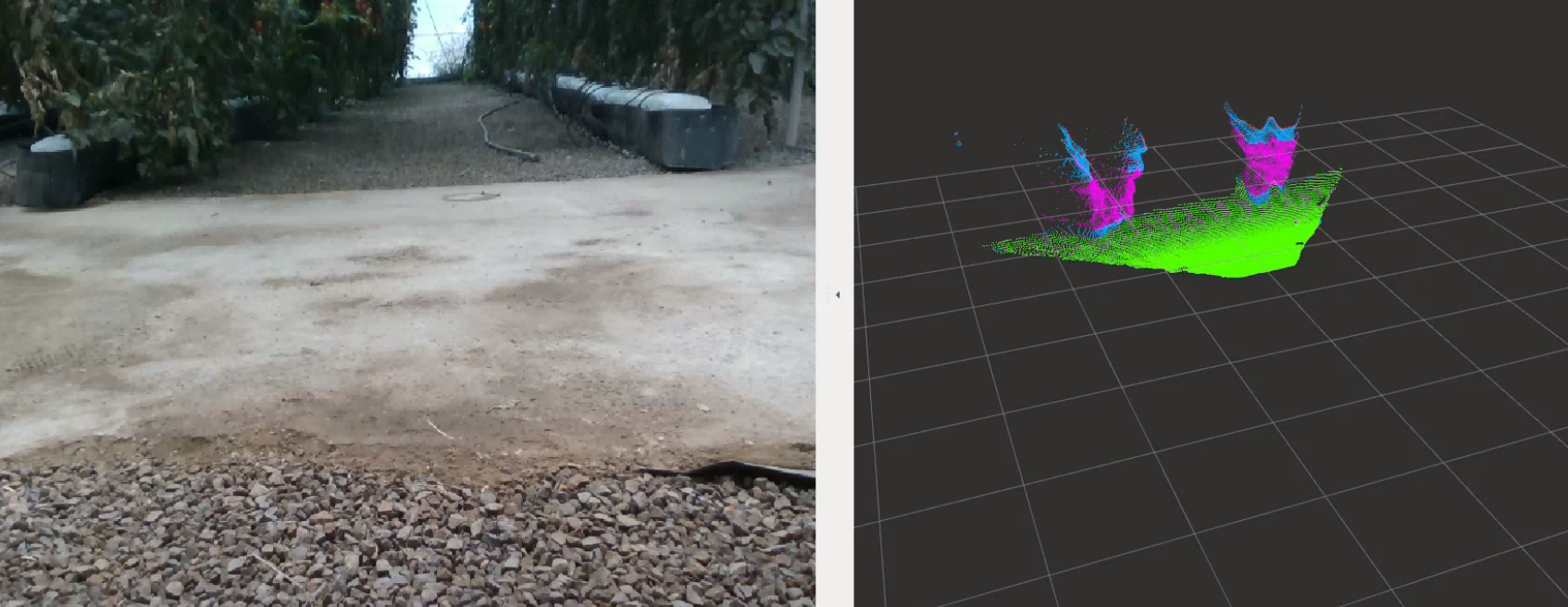}
        }
    \caption{Changing to the next corridor test at four solar conditions. Green points indicates the ground, purple indicates obstacles, and cyan indicates points are higher than the robot}
    \label{fig:4_}
\end{figure}

The corridor transition represents a hybrid challenge, as the robot must cross from one crop row to an adjacent aisle via the central corridor. This maneuver requires high perceptual flexibility to handle abrupt terrain transitions—moving from irregular sand/gravel to smooth concrete and back to sand—while simultaneously managing the shift from the shaded canopy to the briefly exposed central path. The visual results for this transition across the four solar scenarios are shown in Figure~\ref{fig:4_}.

The experimental results for this trajectory are highly positive, demonstrating GreenSeg's superior ability to maintain a stable navigable plane during sudden changes in surface reflectance and local geometry. The diurnal analysis highlights the following strengths:

\begin{itemize}
    \item \textit{Scenario 1 (08:00 h) - Figure \ref{fig:4_1}}: The transition occurs under stable lighting. GreenSeg effectively segments the boundary between the aisle and the corridor, ensuring that the height difference at the concrete edge is not misclassified as an obstacle.
    \item \textit{Scenario 2 (10:00 h) - Figure \ref{fig:4_2}}: Despite glare artifacts emerging as the robot enters the central corridor, the geometric consistency check ($\rho_{min}$) allows the system to ignore these specularities, maintaining an $F_{1,ground}$ of 0.841.
    \item \textit{Scenario 3 (12:00 h) - Figure \ref{fig:4_3}}: At peak irradiance, our method maintains an IoU for the ground class, outperforming the baseline by preventing the ''fragmentation´´ of the ground plane that typically occurs during the material shift.
    \item \textit{Scenario 4 (14:00 h) - Figure \ref{fig:4_4}}: Despite the peak solar elevation, the Region Growing algorithm ensures spatial continuity. 
    \end{itemize}
    
Quantitative results summarized in Table~\ref{tab:test_iv} confirm the robustness of the proposed framework:
\begin{table}[!ht]
\centering
\caption{Results between baseline and GreenSeg for Test 4: Changing to the next corridor}
\label{tab:test_iv}
\renewcommand{\arraystretch}{1.2}
\setlength{\tabcolsep}{2pt} % Optimiza el espaciado interno de las celdas

% Definición del tipo de columna flexible centrada
\newcolumntype{C}{>{\centering\arraybackslash}X}

\begin{tabularx}{\textwidth}{@{} l *{8}{C} @{}}
\toprule
\multirow{2}{*}{\textbf{Class} $k$} & \multicolumn{2}{c}{\textbf{$P_k$}} & \multicolumn{2}{c}{\textbf{$R_k$}} & \multicolumn{2}{c}{\textbf{$F_{1,k}$}} & \multicolumn{2}{c}{\textbf{$\text{IoU}_k$}} \\
\cmidrule(lr){2-3} \cmidrule(lr){4-5} \cmidrule(lr){6-7} \cmidrule(lr){8-9}
& Base & Ours & Base & Ours & Base & Ours & Base & Ours \\ 
\midrule
\textit{ground}   & 0.895 & \textbf{0.925} & 0.880 & \textbf{0.910} & 0.887 & \textbf{0.917} & 0.797 & \textbf{0.846} \\
\textit{obstacle} & 0.840 & \textbf{0.880} & 0.855 & \textbf{0.915} & 0.847 & \textbf{0.897} & 0.734 & \textbf{0.813} \\
\textit{above}    & 0.880 & \textbf{0.915} & 0.875 & \textbf{0.905} & 0.877 & \textbf{0.909} & 0.781 & \textbf{0.833} \\
\textit{noise}    & 0.805 & \textbf{0.845} & 0.765 & \textbf{0.800} & 0.784 & \textbf{0.821} & 0.645 & \textbf{0.696} \\
\midrule
\textbf{Mean}     & 0.855 & \textbf{0.891} & 0.843 & \textbf{0.882} & 0.848 & \textbf{0.886} & 0.739 & \textbf{0.797} \\
\midrule
\textbf{Improvement} & \multicolumn{2}{c}{+4.21\%} & \multicolumn{2}{c}{+4.62\%} & \multicolumn{2}{c}{+4.48\%} & \multicolumn{2}{c}{+7.84\%} \\
\bottomrule
\end{tabularx}
\end{table}

This scenario represents a hybrid linear movement but heterogeneous terrain, often including heating pipes or changes in lighting near the greenhouse headers. While the baseline recovers from the turning maneuver, GreenSeg maintains a clear superiority, delivering a +4.62\% increase in mean Recall and a +7.84\% improvement in mean IoU. These results confirm that the algorithm swiftly adapts to abrupt material and structural shifts, providing a resilient semantic classification of the environment as the robot re-enters the dense crop rows.

The following results were drawn within the scope of the test studies:

\begin{enumerate}
    \item The proposed GreenSeg algorithm consistently outperformed the baseline segmentation framework across all experimental trajectories. The most significant gains were observed during critical rotational maneuvers at the end of the corridors, where the system achieved peak improvements of +11.58\% in mean Recall and +19.24\% in mIoU. These results indicate that the integration of RANSAC global plane fitting combined with local region-growing ensures that the boundaries between ground and obstacles are defined with higher geometric fidelity, drastically reducing the risk of collision in narrow-aisle environments.
    
    \item A major contribution of this work is the effective mitigation of optical and structural artifacts inherent in greenhouse sensing. The integration of seed-point-based geometric validation ($p_s$) combined with curvature filtering ($\kappa_{max}$) proved essential in eliminating "ghost points" and false-positive obstacle detections. These artifacts, typically caused by direct solar glare through frontal openings, specular reflections from the polyethylene cladding, or overhanging leaves, are successfully filtered by ensuring that each segmented point maintains spatial and normal consistency with the robot's localized footprint.
    
    \item The algorithm demonstrated exceptional adaptability to the heterogeneous terrains typical of Mediterranean greenhouses, particularly when navigating between crop rows (yielding a +13.52\% mIoU improvement). By utilizing local surface normal estimation and region-growing connectivity, the system maintained high structural consistency across smooth concrete surfaces, irregular gravel, and tilled soil. This ensures the continuity of the navigable plane even during abrupt material shifts, preventing the segmentation from fragmenting when the reflectance properties of the floor change suddenly.

    \item During end-of-aisle turning maneuvers, the algorithm effectively managed the transient noise and depth discontinuities introduced by the platform's rotational motion. Unlike column-wise scanning methods that fail under angular velocity, the use of a spatial connectivity constraint ensures that only points physically linked to the robot's base are classified as ground. This is a critical factor for maintaining a reliable local costmap in areas where unfiltered sunlight creates significant depth saturation at the greenhouse exits.

    \item The results confirm that the deployment of low-cost RGB-D cameras is sufficient for autonomous tasks in complex environments. GreenSeg bridges the gap between low-cost hardware and the reliability required for agricultural robots, demonstrating that intelligent 3D geometric processing can overcome the hardware limitations typically addressed by expensive 3D LiDAR systems, thereby facilitating the cost-effective scalability of agricultural robotics.

\end{enumerate}

\section{Conclusion} \label{sec: conclusion}

This study has addressed the challenge of ground segmentation in Mediterranean greenhouse environments, where variable lighting, high vegetation density, and terrain heterogeneity typically degrade conventional robotic perception systems. The proposed GreenSeg algorithm demonstrated a significant improvement in mitigating the negative impact of lens flare and "ghost points" resulting from direct solar irradiance and reflections on polyethylene cladding. A fundamental aspect of this robustness is the system's effectiveness when operating between crop rows, where lighting conditions shift constantly due to dynamic shadows projected by the plants and irregular light filtration.

By implementing a dual-consistency framework—combining RANSAC plane fitting with a normal-verified region-growing algorithm—the system successfully maintained the structural integrity of the navigable plane even under critical high-dynamic-range conditions and abrupt rotational movements. Experimental trials conducted on diverse surfaces, including smooth concrete, gravel, and tilled soil, validated that the inclusion of local curvature analysis ($\kappa$) allows the AGRICOBIOT II robot to adapt to irregular floor textures without triggering false positives. This advancement is essential for ensuring fluid mobility during complex tasks, such as navigating between dense plants and executing end-of-aisle turns. Ultimately, reaching peak improvements of 19.24\% in mIoU and 11.58\% in mean Recall during critical turning manoeuvres over baseline methods underscores that the proposed framework provides a resilient and economically viable perception foundation for Mediterranean agriculture. Therefore, the use of this type of low-cost hardware, when paired with robust geometric filtering, constitutes a pivotal contribution to mobile robotics, rivalling the reliability of expensive 3D LiDARs.

As a primary avenue for future research, the integration of the GreenSeg segmentation output directly into an autonomous navigation framework as a dynamic local costmap is envisioned. This approach will facilitate the establishment of a complete and robust autonomous navigation system in Mediterranean greenhouses—a solution based on low-cost hardware that is currently unavailable due to the extreme structural and lighting complexities of these facilities. The ability to transform noisy point clouds between crops into reliable spatial representations for path planning will represent a milestone in the digitization of agricultural tasks, enabling fully autonomous operations in environments where traditional systems consistently fail.

\section*{Acknowledgments} 
The first author, Fernando Cañadas-Aránega, holds an FPI grant (PRE20\\22-102415) from the Spanish Ministry of Science, Innovation, and Universities, and this work was partially supported by Agencia Estatal de Investigación (AEI) under the Project PID2022-139187OB-I00.
\bibliographystyle{cas-model2-names}

\bibliography{cas-refs}




\end{document}